\documentclass[10pt,journal]{IEEEtran}
\usepackage{xcolor}
\usepackage[T1]{fontenc}
\usepackage{times}
\usepackage{soul}
\usepackage{url}
\usepackage[utf8]{inputenc}
\usepackage[small]{caption}
\usepackage{amsmath,amsfonts,amssymb}
\usepackage{algorithmic}
\usepackage{algorithm}
\usepackage{array}
\usepackage[caption=false,font=normalsize,labelfont=sf,textfont=sf]{subfig}
\usepackage{textcomp}
\usepackage{stfloats}
\usepackage{verbatim}
\usepackage{graphicx}
\usepackage{cite}

\usepackage{svg}

\usepackage{caption}

\usepackage{multirow}
\usepackage{booktabs}
\usepackage{ragged2e}
\usepackage{tabularx}
\usepackage{makecell}
\usepackage{wrapfig}
\usepackage{tikz}
\usepackage{comment}
\usepackage[hidelinks]{hyperref}
\hypersetup{
    colorlinks=false,
    pdfborder={0 0 0}
}

\begin{document}

\title{Selective Generative Restoration for Video Deraining via Uncertainty-Aware Diffusion}
 \author{Yi\textquoteleft{}ang~Chen,
         Han~Li,
         Laibin~Chang,
         Hui~Liu,\\
         Kui~Jiang,~\IEEEmembership{Member,~IEEE,}
         Junjun~Jiang,~\IEEEmembership{Senior Member,~IEEE}
        
\thanks{This research was financially supported by the Fundamental and Interdisciplinary Disciplines Breakthrough Plan of the Ministry of Education of China (JYB2025XDXM906) , National Natural Science Foundation of China (62501189) and Harbin Institute of Technology Kunpeng \& Ascend Center of Cultivation. (\emph{Corresponding author: Kui~Jiang})}
\thanks{Yi\textquoteleft{}ang~Chen, Han~Li, Kui~Jiang, and Junjun~Jiang are with the School of Computer Science and Technology, Harbin Institute of Technology, Harbin, China (e-mail: chenya@stu.hit.edu.cn; HYLMW.Li@stu.hit.edu.cn; jiangkui@hit.edu.cn; jiangjunjun@hit.edu.cn).}
\thanks{Laibin~Chang is with the School of Computer and Artificial Intelligence, Zhengzhou University, Zhengzhou 450001, China (e-mail: changlb@zzu.edu.cn).}
\thanks{Hui~Liu is with the Meituan Group, Shanghai 200000, China (e-mail: liuhui132@meituan.com).}
}

\markboth{IEEE Transactions on Multimedia}%
{Chen \MakeLowercase{\textit{et al.}}: SGVD: Selective Generative Restoration for Video Deraining}

\IEEEtitleabstractindextext{
\begin{abstract}
Video deraining is a challenging restoration problem due to complex degradations caused by rain streaks, occlusions, and atmospheric effects.
Existing methods mainly follow a deterministic restoration paradigm, which assumes that the underlying clean scene can be uniquely recovered from rainy observations. However, rain degradation is spatially heterogeneous: reliable regions require faithful reconstruction, whereas severely occluded regions contain ambiguous contents that require stronger prior-driven inference.
Therefore, conventional deterministic restoration often produces over-smoothed results, while global generative restoration may introduce unnecessary hallucinations. In this paper, we introduce a new \textbf{Selective Generative Restoration} paradigm for video deraining, where generative priors are adaptively activated according to restoration uncertainty.
Based on this principle, we propose \textbf{Selective Generative Video Deraining (SGVD)}, an uncertainty-aware diffusion framework that enables adaptive fidelity preservation and generative recovery.
Specifically, SGVD exploits stochastic diversity between dual one-step diffusion predictions to estimate region-level restoration uncertainty, which determines where observation fidelity should be preserved and where generative inference should be activated.
Furthermore, an anchor-guided trajectory consistency mechanism is proposed to regularize stochastic diffusion evolution across frames, reducing temporal generation drift and improving video coherence.
To achieve practical efficiency, SGVD adopts one-step diffusion inference while maintaining uncertainty-aware generative capability.
Extensive experiments demonstrate that SGVD achieves superior restoration performance, perceptual quality, and temporal consistency compared with state-of-the-art video deraining methods, validating the effectiveness of selective generative restoration for complex rainy video restoration.
The source code is available at \url{https://github.com/BrewieCya/SGVD}.

\end{abstract}

}

\maketitle

\IEEEdisplaynontitleabstractindextext

\IEEEpeerreviewmaketitle

\section{Introduction}
\label{sec:introduction}

\begin{figure}[!htb]
\centering
\includegraphics[width=\linewidth]{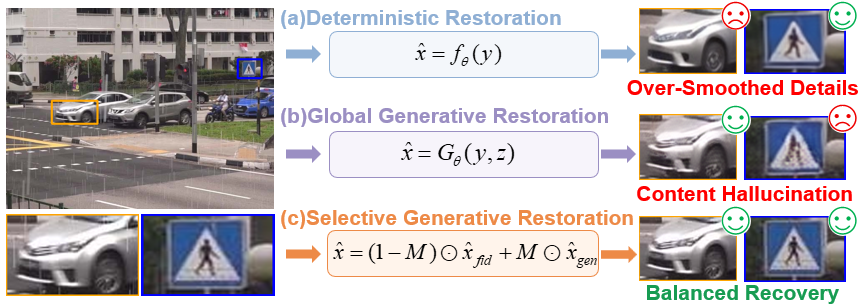}\vspace{-2mm}
\caption{Motivation of Selective Generative Restoration.
Deterministic restoration tends to produce over-smoothed details in severely degraded regions, while global generative restoration may introduce content hallucination in reliable regions.
In contrast, the proposed paradigm performs region-adaptive restoration by preserving fidelity in low-uncertainty regions and selectively activating generative recovery in high-uncertainty regions.}
\vspace{-4mm}
\label{motivation}
\end{figure}

\IEEEPARstart{O}{utdoor} videos captured under rainy conditions often suffer from severe degradation caused by rain streaks, raindrop occlusions, rain accumulation, and atmospheric veiling effects.
These degradations destroy scene structures and temporal consistency, significantly affecting downstream vision tasks such as detection~\cite{li2025domain}, segmentation~\cite{pan2025exploring}, tracking~\cite{liu2024rainytrack}, and autonomous driving~\cite{zhong2022rainy}.
Therefore, robust video deraining is essential for reliable perception under adverse weather conditions.

Existing video deraining methods have achieved remarkable progress through convolutional networks~\cite{li2018video, chen2018robust}, recurrent architectures~\cite{yang2019frame,yang2021recurrent}, Transformers~\cite{wu2023mask,sun2026delivr}, and state space models~\cite{wu2024rainmamba, sun2025semi}, mainly by improving spatial representation and temporal dependency modeling.
However, most methods follow a deterministic restoration paradigm that maps each rainy observation to a single clean output.
Such a formulation assumes that the underlying scene can be uniquely recovered from degraded observations, which becomes problematic under severe rain occlusions where background information is partially missing and multiple plausible restorations may exist.

Unlike uniformly degraded restoration problems, rain degradation is inherently spatially heterogeneous. Reliable regions preserve sufficient observations and mainly require fidelity-oriented reconstruction, whereas heavily occluded regions contain uncertain contents and require stronger prior-driven inference.
Therefore, video deraining is not merely a reconstruction problem, but a spatially varying inverse problem with different levels of restoration uncertainty. This observation reveals a fundamental limitation of existing restoration paradigms. Global deterministic restoration~\cite{liao2024asymmetric,liu2024fastglobal,liu2022generalizedsmoothing} tends to over-smooth ambiguous regions, while global generative restoration may introduce unnecessary hallucinations into reliable regions. Consequently, effective video deraining requires an adaptive strategy that determines where observation fidelity should dominate and where generative priors should be activated.

Recent diffusion models provide a promising direction by modeling conditional distributions rather than predicting a single deterministic solution.
Their powerful generative priors make them suitable for recovering missing contents from ambiguous observations.
However, directly applying diffusion models to video deraining remains challenging because rain degradation exhibits strong spatial uncertainty variation.
As illustrated in Fig.~\ref{motivation}, uniformly applying diffusion priors may improve uncertain regions but simultaneously alter trustworthy contents.
Thus, the key problem is not whether diffusion should be introduced, but how to selectively activate generation according to restoration uncertainty.

To address this issue, we introduce a new restoration paradigm, termed
\textbf{Selective Generative Restoration}.
The core idea is that diffusion priors should be adaptively activated according to region-level uncertainty:
low-uncertainty regions preserve observation fidelity, while high-uncertainty regions exploit generative inference for plausible recovery.

However, realizing Selective Generative Restoration requires addressing three challenges.
First, an explicit uncertainty estimation mechanism is required to determine where generative restoration is necessary.
Second, stochastic diffusion prediction may introduce inconsistent restoration trajectories across frames, causing temporal flickering.
Third, conventional iterative diffusion sampling is computationally expensive and limits practical video restoration applications.

To address these challenges, we propose
\textbf{Selective Generative Video Deraining (SGVD)},
an uncertainty-aware diffusion framework supporting one-step restoration, built upon the principle of Selective Generative Restoration. SGVD uses stochastic prediction disagreement as a practical signal of restoration uncertainty.
Specifically, given the same rainy observation, SGVD performs two one-step diffusion predictions with different noise perturbations and measures their latent discrepancy as an approximation of posterior uncertainty.
The resulting uncertainty map explicitly identifies regions with different restoration difficulty: low-uncertainty regions contain reliable observations and are reconstructed with stronger fidelity constraints, whereas high-uncertainty regions contain missing or ambiguous contents and are enhanced through selective generative inference.

Furthermore, to address temporal instability caused by stochastic diffusion prediction, SGVD introduces an anchor-guided trajectory consistency mechanism.
The proposed strategy regularizes relative trajectory distances to an anchor obtained by averaging predicted latent representations within a five-frame subclip. This local temporal constraint is applied during training with 49-frame inputs.
This design reduces frame-wise generation drift and encourages temporally coherent restoration across neighboring frames. Finally, to achieve efficient diffusion-based video deraining, SGVD adopts a one-step inference strategy that directly predicts clean latent representations from rainy inputs.
Different from conventional iterative diffusion restoration, the proposed formulation preserves the uncertainty-aware generative capability while substantially reducing sampling complexity, making diffusion-based video deraining more practical for real-world applications.

Overall, SGVD establishes a new perspective that diffusion restoration should move beyond global generation toward uncertainty-aware selective generation.
By adaptively balancing observation fidelity, generative inference, and temporal consistency, SGVD provides an effective solution for restoring rainy videos with heterogeneous degradation uncertainty.

Overall, the main contributions of this work are summarized as follows.
\begin{itemize}
    \item We introduce a new \textbf{Selective Generative Restoration} paradigm for video deraining, revealing that rain degradation is a spatially heterogeneous inverse problem where different regions require different balances between observation fidelity and generative inference. %

    \item We propose \textbf{SGVD}, a one-step uncertainty-aware diffusion framework that selectively activates generative restoration according to region-level posterior uncertainty. By exploiting stochastic diversity between dual diffusion predictions, SGVD constructs an explicit uncertainty representation and achieves adaptive restoration between faithful reconstruction and plausible content recovery.

    \item We develop an anchor-guided trajectory consistency mechanism to regularize stochastic diffusion evolution across video frames. Together with the proposed one-step inference, SGVD effectively suppresses temporal generation drift while maintaining high restoration efficiency, enabling practical diffusion-based video deraining.

\end{itemize}

\section{RELATED WORK}
\label{sec:relat}

\subsection{Video Deraining}
Video deraining has evolved from prior-based rain--background decomposition to learned spatio-temporal restoration. Early methods exploited rain sparsity and background low-rankness~\cite{kim2015video,ren2017video}, while subsequent CNN and recurrent models learned spatial rain representations and temporal dependencies directly from data~\cite{chen2018robust,liu2018d3r,yang2021recurrent}. Transformer-based methods further improved long-range interaction and motion-aware aggregation~\cite{wen2023video,sun2026delivr}, and recent state-space models enhanced long-sequence modeling efficiency~\cite{wu2024rainmamba,sun2025semi}. Despite these advances, most methods formulate video deraining as deterministic point estimation. This formulation is effective when sufficient background evidence remains observable, but often produces over-smoothed results in severely occluded regions where multiple plausible restorations may exist. Recent works have therefore explored pretrained diffusion priors for video deraining, including zero-shot restoration~\cite{varanka2026zsvd} and nighttime rain removal~\cite{yang2026uenr}. However, existing generative approaches generally apply diffusion priors globally and do not explicitly distinguish reliable observations from ambiguous rain-occluded regions.

\begin{figure*}[!htb]
\centering
\includegraphics[width=0.92\textwidth]{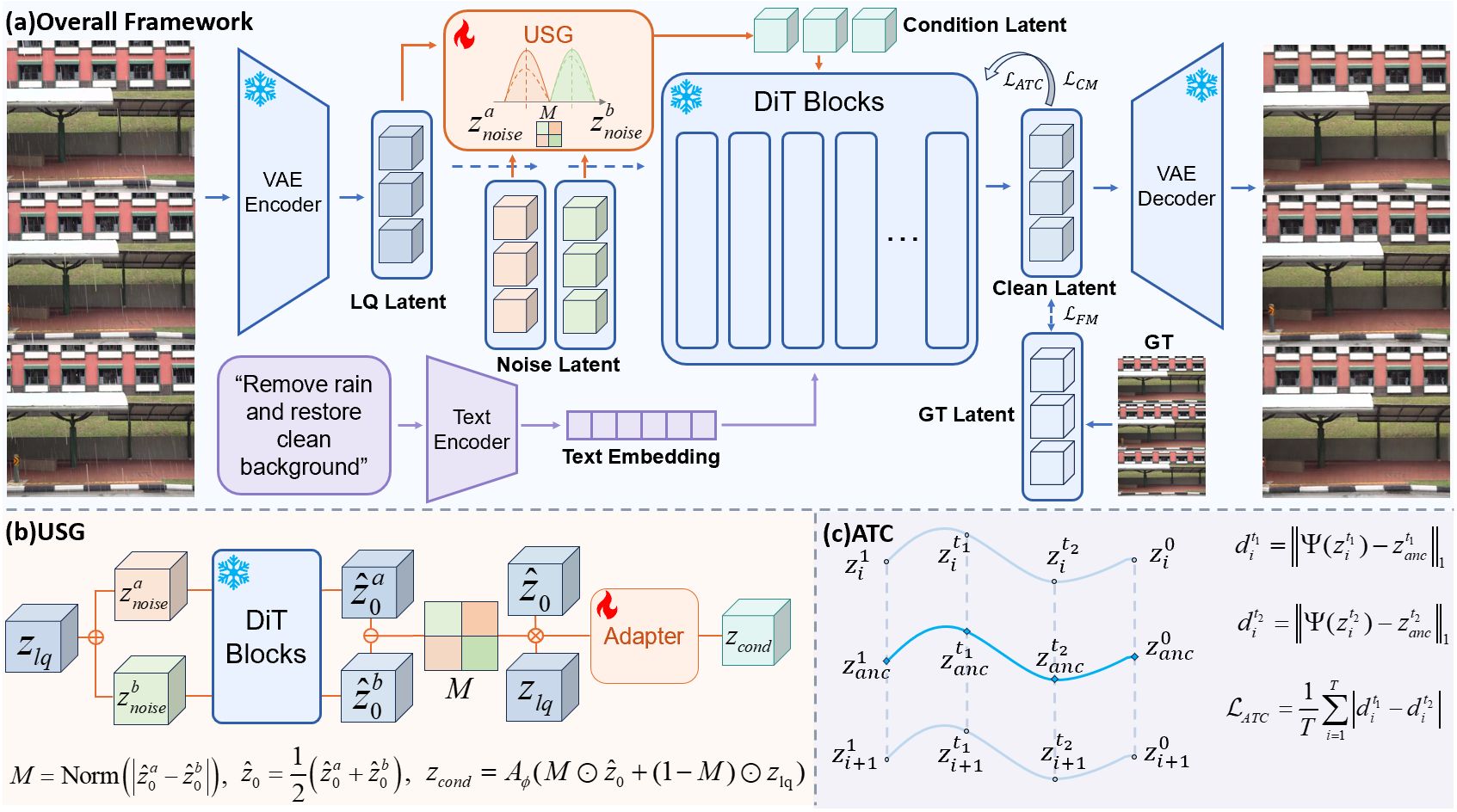}\vspace{-2mm}
\captionsetup{font={scriptsize}}
\caption{Overall architecture of the proposed SGVD for video deraining.
The rainy video is first encoded into the latent space by a frozen VAE encoder.
The Uncertainty-Guided Selective Generation (USG) module estimates a spatial uncertainty map from the discrepancy between dual-noise clean latent predictions, and adaptively combines the generative latent with the fidelity-preserving LQ latent.
The resulting conditional latent is injected into the frozen DiT backbone through a lightweight trainable adapter.
Meanwhile, Anchor-Guided Trajectory Consistency (ATC) regularizes each frame\textquotesingle s distance to a shared five-frame anchor across trajectory states, promoting temporal coherence without evaluating this training loss at inference.}
\label{Framework}
\vspace{-4mm}
\end{figure*}

\subsection{Diffusion Models for Low-Level Vision}
Diffusion models extend low-level vision from deterministic prediction~\cite{zhu2026isfm,sun2025spatial} to conditional distribution modeling, enabling plausible recovery when observations are incomplete. SR3~\cite{saharia2023sr3} introduced conditional diffusion for super-resolution, while StableSR~\cite{wang2024stablesr} leveraged pretrained latent diffusion priors for realistic texture reconstruction. Subsequent studies have applied diffusion models to various image and video restoration tasks~\cite{cai2026coen,liu2025ultravsr,gu2026consis,han2025dcvsr}. Nevertheless, strong generative priors introduce a fidelity--perception trade-off. FiDeSR~\cite{kim2026fidesr} regulates this trade-off for one-step super-resolution, and WeatherDiffusion~\cite{weatherdiffusion} supports weather-guided forward and inverse rendering. These methods mainly control generation globally, whereas rain degradation is spatially heterogeneous. In contrast, SGVD estimates region-level posterior uncertainty from stochastic prediction diversity and uses it to selectively preserve reliable observations or activate generative recovery.

\subsection{Efficient Generative Models}
The iterative sampling process of diffusion models limits their applicability to video restoration. DDIM~\cite{song2021ddim} and DPM-Solver~\cite{lu2022dpm} accelerate inference through shortened or improved numerical sampling. Flow Matching~\cite{lipman2023flow} and Rectified Flow~\cite{liu2022flow} instead learn continuous transport trajectories and have supported scalable video generators such as Wan~\cite{wan2025wan}. Consistency Models~\cite{song2023consistency} and Latent Consistency Models~\cite{lu2023latent} further compress multi-step trajectories into few-step or one-step prediction, with SinSR~\cite{wang2024sinsr} and OSEDiff~\cite{wu2024osediff} demonstrating efficient one-step image restoration. However, sampling acceleration alone does not ensure temporally stable video restoration, since stochastic predictions of neighboring frames may follow inconsistent generative trajectories. SGVD combines one-step generative inference with Anchor-Guided Trajectory Consistency, which regularizes relative inter-frame evolution during training without additional inference cost.

\section{Selective Generative Restoration Framework}
\label{sec:Method}

\subsection{Overview of SGVD}
\label{sec:overview}
The proposed SGVD is built upon the principle of \textbf{Selective Generative Restoration}, which aims to adaptively balance observation fidelity and generative inference according to restoration uncertainty. Different from conventional restoration frameworks that apply a uniform reconstruction strategy to all regions, SGVD considers rainy video restoration as a spatially heterogeneous inverse problem.
Specifically, regions with reliable observations should preserve the original visual evidence, while regions with severe information loss should exploit generative priors to recover plausible contents.

Given a rainy video sequence $\mathbf{X}=\{x_i\}_{i=1}^{T},
$ SGVD targets the recovery of a clean and temporally coherent video
$
\hat{\mathbf{Y}}=\{\hat{y}_i\}_{i=1}^{T}.
$ As illustrated in Fig.~\ref{Framework}, SGVD consists of three key components:
(1) a \textbf{One-Step Generative Restoration Backbone} that provides efficient latent-space generative reconstruction;
(2) an \textbf{Uncertainty-Guided Selective Generation (USG)} module that estimates region-level restoration uncertainty and adaptively controls the activation of generative priors;
and (3) an \textbf{Anchor-Guided Trajectory Consistency (ATC)} mechanism that regularizes stochastic generation trajectories to maintain temporal coherence.

Specifically, the rainy video is first encoded into the latent space using a frozen VAE encoder.
The latent representation is then processed by the USG module, where stochastic generative predictions are exploited to estimate restoration uncertainty.
Based on the estimated uncertainty, SGVD selectively integrates fidelity-preserving rainy latent features and generative clean latent predictions to construct an adaptive restoration condition.

The resulting condition and uncertainty map are injected into the frozen generative backbone through a trainable adapter. We use the fixed prompt \emph{Remove rain and restore clean background} during training and standard testing. The frozen text encoder supplies the task instruction, while the rainy video and uncertainty-guided latent condition provide scene-specific structure and appearance.
During training, ATC regularizes the relative evolution of neighboring frames along the generative trajectory, suppressing stochastic temporal drift without additional inference cost. Together with uncertainty-guided selective generation, SGVD adaptively balances fidelity preservation and generative recovery for complex rainy videos.

\subsection{One-Step Generative Restoration Backbone}
\label{sec:backbone}

Selective Generative Restoration requires a generative model that can provide strong prior knowledge while maintaining practical restoration efficiency.
However, conventional diffusion models typically initialize generation from pure Gaussian noise and rely on iterative denoising, which may introduce excessive generation freedom and substantial computational overhead for video restoration.
SGVD uses the pretrained Wan2.1-T2V-1.3B backbone~\cite{wan2025wan} with an LQ-aware restoration condition. Flow matching and consistency learning provide the training formulation for predicting clean latent endpoints.

Given a GT latent representation $z_0$, an LQ latent representation $z^{lq}_1$ and a Gaussian noise latent $z^{noise}_1\sim\mathcal{N}(0,\mathbf{I})$, we first construct an LQ-aware noisy latent:
\begin{equation}
z_1
=
\alpha z^{lq}_1
+
\sigma z^{noise}_1,
\end{equation}
where $\alpha$ and $\sigma$ control the contributions of the LQ observation and stochastic perturbation, respectively.
Then flow matching defines a continuous transport trajectory between the data distribution and noise distribution:
\begin{equation}
z_t=(1-t)z_0+t z_1,
\qquad t\in[0,1].
\end{equation}
The corresponding velocity field is
$u_t=
\frac{d z_t}{d t}
=
z_1-z_0 $.
A conditional velocity network $v_{\theta,\phi}$ approximates this transport direction, where $\theta$ denotes the frozen pretrained parameters and $\phi$ denotes the trainable adapter parameters.
\begin{equation}
\mathcal{L}_{\mathrm{FM}}
=
\mathbb{E}_{t,z_0,z_1}
\left[
\left\|
v_{\theta,\phi}(z_t,t,c)-u_t
\right\|_2^2
\right],
\end{equation}
where $c$ denotes the condition extracted from the rainy video and text embedding.

Although flow matching provides an efficient formulation for generative modeling, directly solving the continuous trajectory still requires numerical integration.
Therefore, SGVD further adopts consistency learning following Consistency Models~\cite{song2023consistency} to compress the generative trajectory into a direct endpoint prediction.

The consistency function $F_{\theta,\phi}$ is trained to produce consistent clean latent predictions from different states along the same trajectory.
\begin{equation}
F_{\theta,\phi}(z_t,t,c)
=
F_{\theta,\phi}(z_s,s,c)
=
z_0 ,
\end{equation}
where $s<t$. The corresponding consistency objective is formulated as
\begin{equation}
\mathcal{L}_{\mathrm{CM}}
=
\left\|
F_{\theta,\phi}(z_t,t,c)
-
\mathrm{sg}
\left[
F_{\theta,\bar{\phi}}(z_s,s,c)
\right]
\right\|_2^2 ,
\end{equation}
Here $\bar{\phi}$ denotes the exponential moving average of the adapter parameters, while the pretrained weights $\theta$ remain fixed. The operator $\mathrm{sg}(\cdot)$ stops gradients through the consistency target.

\subsection{Diffusion Posterior Uncertainty Estimation}
\label{sec:uncertainty}

Selective Generative Restoration requires identifying which regions should preserve observed evidence and which regions require generative inference.
Therefore, estimating region-level restoration uncertainty is the fundamental step for adaptive generation. Given a rainy video sequence $\mathbf{X}$, the frozen VAE encoder first maps the input into the latent space: $z^{lq}_1=\mathcal{E}(\mathbf{X})$, where $z^{lq}_1$ represents the latent observation containing both reliable scene information and rain-induced degradation. Unlike deterministic restoration models that estimate a single solution, generative models characterize a conditional distribution: $p(z_0|z^{lq}_1)$, where $z_0$ denotes the latent representation of possible clean videos.
The uncertainty of this posterior reflects the ambiguity of restoration:
regions with sufficient observations exhibit concentrated predictions, whereas severely occluded regions produce diverse plausible solutions.

To approximate this posterior uncertainty, SGVD exploits the stochasticity of the generative trajectory.
Specifically, two independent noise perturbations are injected into the same rainy latent representation:
\begin{equation}
z_t^{a}
=
\alpha_t z^{lq}_1
+
\sigma_t\epsilon^{a},
~~z_t^{b}
=
\alpha_t z^{lq}_1
+
\sigma_t\epsilon^{b},
\end{equation}
where
$\epsilon^{a},\epsilon^{b}\sim\mathcal{N}(0,\mathbf{I})$. The two perturbed states are processed by the frozen generative backbone without the adapter.

\begin{equation}
\hat{v}^{a}
=
\mathcal{F}^{\mathrm{frz}}
(z_t^{a},t,z^{lq}_1),
~~\hat{v}^{b}
=
\mathcal{F}^{\mathrm{frz}}
(z_t^{b},t,z^{lq}_1).
\end{equation}
Based on the predicted velocity fields, the corresponding clean latent estimations are recovered as:
\begin{equation}
\hat{z}_{0}^{a}
=
z_t^{a}
-
t\hat{v}^{a},
~~\hat{z}_{0}^{b}
=
z_t^{b}
-
t\hat{v}^{b}.
\end{equation}

Although both predictions are generated from the same rainy observation, their divergence reflects the uncertainty of the conditional restoration distribution.
Specifically, low-uncertainty regions with reliable visual evidence produce consistent clean predictions, while ambiguous regions with missing background information exhibit larger stochastic variations. Accordingly, SGVD estimates the diffusion posterior uncertainty map as:
\begin{equation}
M
=
\mathcal{N}
\left(
\operatorname{Mean}_{C}
(
|
\hat{z}_{0}^{a}
-
\hat{z}_{0}^{b}
|
)
\right).
\end{equation}
Here $\operatorname{Mean}_{C}$ averages channels, and $\mathcal N$ performs $3\times3$ spatial mean filtering without temporal smoothing, followed by clip-wise min--max normalization.

The obtained uncertainty map $M\in[0,1]$ provides an explicit spatial measurement of restoration ambiguity:
larger values indicate regions with higher posterior uncertainty and stronger demand for generative recovery, while smaller values correspond to reliable regions where observation fidelity should be maintained.

\subsection{Uncertainty-Guided Selective Generation}
\label{sec:usg}
The estimated posterior uncertainty provides a spatial criterion for selectively activating generative restoration.
Instead of applying generative priors uniformly, SGVD dynamically balances observation fidelity and generative inference according to the uncertainty map. First, the two stochastic clean latent predictions are aggregated to obtain a generative restoration proposal:
$\hat{z}_{0}
=
\frac{1}{2}
(
\hat{z}_{0}^{a}
+
\hat{z}_{0}^{b}
)$.
To alleviate the unnecessary hallucinations in reliable regions, SGVD thus performs uncertainty-guided selective fusion between the generative prediction and the observed rainy latent representation:
\begin{equation}
z_{\mathrm{sel}}
=
M\odot \hat{z}_{0}
+
(1-M)\odot z^{lq}_1,
\end{equation}
where $\odot$ denotes element-wise multiplication. This formulation explicitly realizes the principle of Selective Generative Restoration.
For high-uncertainty regions,
$M$ increases and activates the generative prior to recover ambiguous contents.
For low-uncertainty regions,
$(1-M)$ dominates and preserves trustworthy observations.

To adapt the selected latent representation to the frozen generative backbone, a lightweight trainable adapter is introduced:
$z_{\mathrm{cond}}=\mathcal{A}_{\phi}(z_{\mathrm{sel}},M)$,
where $\mathcal{A}_{\phi}$ denotes the trainable adapter. It uses a convolutional embedding module to map the concatenated $z_{\mathrm{sel}}$ and $M$ into tokens aligned with the DiT grid, followed by block-specific linear projections. The resulting condition is injected into the generative backbone by scaling the projected features with pooled uncertainty and adding them to each DiT block's output to produce the restored clean latent representation. The injection projections are initialized to zero to preserve the backbone's initial behavior. Through uncertainty-guided selective activation, SGVD avoids the two opposite failure modes of previous restoration paradigms:
over-smoothing caused by excessive fidelity constraints and hallucination caused by indiscriminate generation.

Therefore, USG transforms diffusion generation from a global restoration process into an uncertainty-aware selective restoration process, enabling adaptive recovery for spatially heterogeneous rainy videos.

\subsection{Anchor-Guided Trajectory Consistency}
\label{sec:atc}
Although uncertainty-guided selective generation enables adaptive activation of generative priors, stochastic generation may still introduce temporal trajectory divergence among consecutive frames.
Different from deterministic restoration, where each frame is mapped to a fixed output, generative restoration produces a distribution of possible restoration trajectories.
Therefore, maintaining temporal consistency requires regularizing the evolution of generative trajectories rather than only constraining the final restored frames.

To address this issue, we introduce an \textbf{Anchor-Guided Trajectory Consistency (ATC)} mechanism. The key idea is to construct a shared subclip-level temporal anchor and enforce the relative trajectory evolution of neighboring frames to remain consistent during one-step generative restoration. The shared anchor provides a common reference that incorporates all frames within a subclip, offering broader temporal context than an immediate-neighbor constraint. Given a five-frame subclip from a 49-frame training input $\mathbf{z}=\{z_i\}_{i=1}^{K}$ with $K=5$, the generative backbone predicts clean latent representations at different trajectory states:
\begin{equation}
\hat{z}_{i}^{t}
=
\mathcal{F}_{\theta,\phi}
(z_i^{t},t,z_{\mathrm{cond}}),
\end{equation}
where $t$ denotes the flow trajectory state and $z_{\mathrm{cond}}$ represents the uncertainty-guided selective condition.

To provide a stable temporal reference, we construct a subclip-level anchor representation:
\begin{equation}
z^t_{\mathrm{anc}}
=
\frac{1}{K}
\sum_{i=1}^{K}
\Psi(\hat{z}_{i}^{t}),
\end{equation}
where $\Psi(\cdot)$ denotes a latent projection function. We use the parameter-free identity mapping $\Psi(z)=z$ for all frames and trajectory states.

The anchor captures the shared semantic structure of the subclip while preserving frame-specific variations.
Based on this anchor, we measure the relative distance between each frame trajectory and the shared subclip-level reference:
\begin{equation}
d_i^{t}
=
\left\|
\Psi(\hat{z}_{i}^{t})
-
z^t_{\mathrm{anc}}
\right\|_1 .
\end{equation}

For stochastic generative trajectories at two different states $t_1$ and $t_2$, temporally coherent restoration should maintain consistent relative distances: $d_i^{t_1}
\approx
d_i^{t_2}$. Accordingly, the proposed ATC objective is formulated as:
\begin{equation}
\mathcal{L}_{\mathrm{ATC}}
=
\frac{1}{K}
\sum_{i=1}^{K}
\left|
d_i^{t_1}
-
d_i^{t_2}
\right|.
\end{equation}

ATC regularizes changes in each frame's distance to the shared anchor across trajectory states, allowing frame-specific motion and appearance while discouraging stochastic temporal drift. By introducing anchor-guided trajectory regularization, SGVD extends selective generation from spatial adaptation to temporally consistent generative restoration, enabling stable video recovery under stochastic inference.
\subsection{Optimization Objective}
\label{sec:optimization}

During training, SGVD optimizes the selective generative restoration process while keeping the pretrained generative prior fixed.
Only the adapter parameters $\phi$ are trainable. The pretrained DiT backbone, VAE encoder and decoder, and text encoder remain frozen.
The overall training objective consists of three complementary terms:
(1) a flow matching objective for learning the generative transport trajectory,
(2) a consistency objective for compressing the continuous trajectory into one-step restoration,
and (3) the proposed anchor-guided trajectory consistency objective for temporally coherent generation.
\begin{table*}[!htbp]
\centering
\captionsetup{font={scriptsize}}
\caption{Quantitative comparisons with representative video restoration and task-specific video deraining methods on synthetic and real-world video deraining datasets. The best and second-best results are highlighted in \textbf{bold} and \underline{underline}, respectively.}
\vspace{-2mm}
\label{tab:sota}
\setlength{\tabcolsep}{2.5pt}
\renewcommand{\arraystretch}{1.05}
\resizebox{0.96\textwidth}{!}{
\begin{tabular}{ll|cccccccccc}
\toprule
Dataset
& Metric
& DiffIR2VR~\cite{yeh2024diffir2vrzero}
& Diff-Plugin~\cite{liu2024diffplugin}
& T3-Diff~\cite{chen2024t3diffweather}
& WeatherDiff~\cite{weatherdiffusion}
& DiffBIR~\cite{lin2023diffbir}
& ZSVD~\cite{varanka2026zsvd}
& DOVE~\cite{dove}
& RainMamba~\cite{wu2024rainmamba}
& DeLiVR~\cite{sun2026delivr}
& \textbf{SGVD} \\
\midrule

\multirow{7}{*}{NTURain-Syn}

& PSNR$\uparrow$
& 23.7304
& 24.1138
& 25.3339
& 25.4756
& 25.9601
& 25.9677
& 26.2488
& \underline{36.3430}
& \textbf{36.7115}
& 26.5174
\\

& SSIM$\uparrow$
& 0.6850
& 0.7161
& 0.8633
& 0.8110
& 0.7802
& 0.8027
& 0.8534
& \textbf{0.9705}
& \underline{0.9684}
& 0.8810
\\

& LPIPS$\downarrow$
& 0.2755
& 0.1137
& 0.1350
& 0.1471
& 0.1221
& 0.1937
& 0.1333
& \textbf{0.0257}
& \underline{0.0281}
& 0.1310
\\

& NIQE$\downarrow$
& 4.2206
& 3.8413
& 3.4850
& \underline{3.4752}
& 3.6846
& 3.7548
& 4.8571
& 3.6070
& 3.5866
& \textbf{3.4224}
\\

& CLIP-IQA$\uparrow$
& \textbf{0.3601}
& 0.3273
& 0.2877
& 0.2918
& 0.2583
& 0.2452
& 0.2537
& 0.2766
& 0.2666
& \underline{0.3565}
\\

& MUSIQ$\uparrow$
& 57.3385
& \textbf{59.4466}
& 57.1062
& 58.2231
& 53.3840
& 53.1633
& 52.7594
& 57.2592
& 56.4605
& \underline{58.3749}
\\

& WE$\downarrow$
& 0.0286
& 0.0351
& \underline{0.0210}
& 0.0286
& 0.0249
& 0.0226
& 0.0241
& 0.0214
& 0.0221
& \textbf{0.0207}
\\

\midrule

\multirow{7}{*}{RainSynLight25}

& PSNR$\uparrow$
& 22.7688
& 21.9237
& 24.8369
& 24.3363
& 23.7407
& 21.7876
& \underline{25.5674}
& 24.2891
& 24.3912
& \textbf{25.9165}
\\

& SSIM$\uparrow$
& 0.6314
& 0.6366
& \underline{0.8439}
& 0.8387
& 0.7068
& 0.6393
& 0.7734
& 0.8364
& 0.8382
& \textbf{0.8492}
\\

& LPIPS$\downarrow$
& 0.3657
& \textbf{0.1962}
& 0.2547
& \underline{0.2430}
& 0.2908
& 0.3620
& 0.2664
& 0.3590
& 0.3722
& 0.3013
\\

& NIQE$\downarrow$
& 5.1659
& \textbf{3.7525}
& 4.5336
& 4.5294
& 4.5301
& 4.4077
& 5.1229
& 4.4971
& 4.2732
& \underline{4.1671}
\\

& CLIP-IQA$\uparrow$
& 0.4660
& \underline{0.4837}
& 0.4106
& 0.3943
& 0.3814
& 0.3257
& 0.3513
& 0.4003
& 0.4249
& \textbf{0.4971}
\\

& MUSIQ$\uparrow$
& 58.0105
& 57.5581
& 58.1555
& 57.5304
& 52.6834
& 50.7805
& 54.4247
& \underline{58.2881}
& 58.1040
& \textbf{58.4274}
\\

& WE$\downarrow$
& 0.0503
& 0.0444
& 0.0489
& 0.0527
& 0.0470
& 0.0446
& \textbf{0.0255}
& 0.0444
& 0.0367
& \underline{0.0347}
\\

\midrule

\multirow{7}{*}{RainSynComplex25}

& PSNR$\uparrow$
& 15.2853
& 16.7380
& 17.0794
& 16.0397
& 17.0434
& 15.2853
& 18.5361
& 17.1749
& \textbf{18.7076}
& \underline{18.5545}
\\

& SSIM$\uparrow$
& 0.4172
& 0.4647
& 0.5772
& 0.5267
& 0.5186
& 0.3862
& \underline{0.5810}
& 0.5505
& 0.5432
& \textbf{0.5812}
\\

& LPIPS$\downarrow$
& 0.5225
& \textbf{0.3437}
& 0.4135
& 0.4583
& 0.4244
& 0.5434
& \underline{0.4036}
& 0.5212
& 0.4991
& 0.4922
\\

& NIQE$\downarrow$
& 8.3184
& \textbf{3.8672}
& 5.6912
& 8.0030
& 5.2094
& 6.6324
& 5.6132
& 5.9325
& 5.0987
& \underline{4.2315}
\\

& CLIP-IQA$\uparrow$
& \underline{0.3905}
& \textbf{0.4257}
& 0.3168
& 0.3060
& 0.3211
& 0.2465
& 0.2934
& 0.3029
& 0.3069
& 0.3521
\\

& MUSIQ$\uparrow$
& 56.6605
& \underline{58.1845}
& 56.6112
& 57.8192
& 50.8225
& 43.5738
& 48.5194
& 54.1892
& 53.8462
& \textbf{58.8422}
\\

& WE$\downarrow$
& 0.1180
& \underline{0.0783}
& 0.1239
& 0.1438
& 0.1117
& 0.1304
& \textbf{0.0446}
& 0.1221
& 0.1001
& 0.1014
\\

\midrule

\multirow{4}{*}{NTURain-Real}

& NIQE$\downarrow$
& 4.2686
& 3.7203
& 3.3842
& 3.3784
& 3.9342
& 3.9149
& 4.9289
& 3.3609
& \underline{3.3340}
& \textbf{3.2927}
\\

& CLIP-IQA$\uparrow$
& 0.3385
& \textbf{0.4077}
& 0.3786
& \underline{0.3947}
& 0.3401
& 0.2755
& 0.2360
& 0.3618
& 0.3440
& 0.3930
\\

& MUSIQ$\uparrow$
& 60.2405
& 62.0713
& 62.5504
& \textbf{64.3716}
& 57.2874
& 54.7643
& 53.7124
& 62.2050
& 62.3364
& \underline{62.6503}
\\

& WE$\downarrow$
& 0.0276
& 0.0395
& 0.0335
& 0.0327
& 0.0326
& 0.0258
& \underline{0.0205}
& 0.0325
& 0.0315
& \textbf{0.0204}
\\

\bottomrule
\end{tabular}}
\vspace{-3mm}
\end{table*}

The flow matching objective $\mathcal{L}_{\mathrm{FM}}$ learns the transport direction from LQ-aware noisy latents toward the clean latent distribution, while the consistency objective $\mathcal{L}_{\mathrm{CM}}$ encourages different states along the same trajectory to predict a consistent clean endpoint for efficient one-step restoration. In addition, the proposed $\mathcal{L}_{\mathrm{ATC}}$ regularizes frame-wise latent trajectories relative to the shared anchor of each five-frame subclip, suppressing stochastic temporal drift while preserving natural motion and frame-specific variations.

Combining the above objectives, the final training objective is formulated as:
\begin{equation}
\mathcal{L}_{\mathrm{total}}
=
\lambda_{\mathrm{FM}}
\mathcal{L}_{\mathrm{FM}}
+
\lambda_{\mathrm{CM}}
\mathcal{L}_{\mathrm{CM}}
+
\lambda_{\mathrm{ATC}}
\mathcal{L}_{\mathrm{ATC}},
\end{equation}
where
$\lambda_{\mathrm{FM}}$,
$\lambda_{\mathrm{CM}}$,
and
$\lambda_{\mathrm{ATC}}$
are the balancing coefficients for flow matching, one-step consistency, and anchor-guided trajectory consistency, respectively.

\newcommand{\sgvdplatformnote}{We conducted experiments on two 48~GB 4090 NVIDIA GPUs and also ran on a system with a Kunpeng~920 CPU and an Ascend~910B4 NPU.}
\section{Experimental Results}
\label{sec:EX}

\subsection{Experimental Settings}
\label{sec:ID}
\noindent\textbf{Datasets.}
We evaluate SGVD on NTURain-Syn~\cite{chen2018robust}, RainSynLight25, RainSynComplex25~\cite{liu2018d3r}, and real-world NTURain videos without clean references. The NTURain-Syn training set contains 25 rainy sequences synthesized from eight clean videos with varying rain severity. All methods are retrained on the same NTURain-Syn split and evaluated on common test splits, with direct cross-dataset evaluation on RainSynLight25 and RainSynComplex25 without fine-tuning.

\noindent\textbf{Evaluation Metrics.}
We report PSNR and SSIM for reconstruction fidelity; LPIPS, NIQE, CLIP-IQA, and MUSIQ for perceptual quality; and optical-flow-based warping error (WE) for temporal consistency, using RAFT~\cite{teed2020raft} for flow estimation. All methods share the metric computation protocol. For real-world videos without clean references, we report NIQE, MUSIQ, CLIP-IQA, and WE together with qualitative results.

\begin{figure*}[!htb]
\centering
\captionsetup{font={scriptsize}}

\setlength{\tabcolsep}{1pt}

{\footnotesize
\begin{tabular}{cccccccc}

\includegraphics[width=0.12\textwidth]{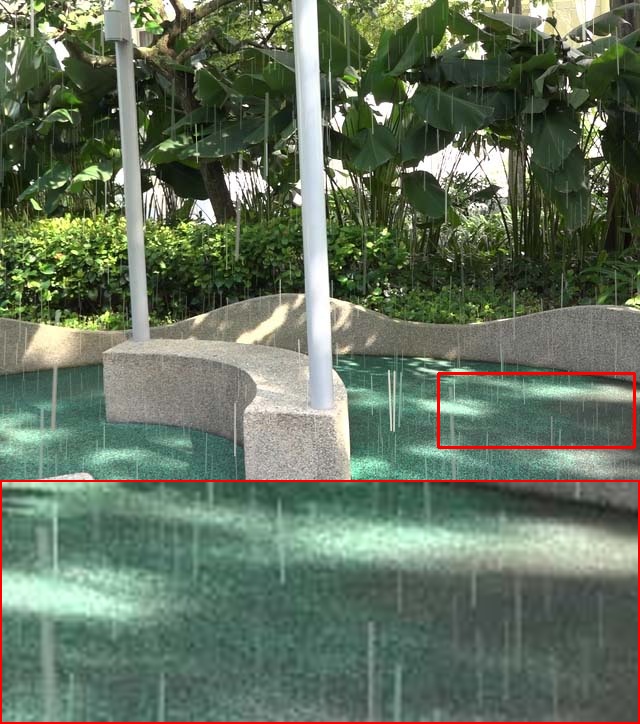}
&
\includegraphics[width=0.12\textwidth]{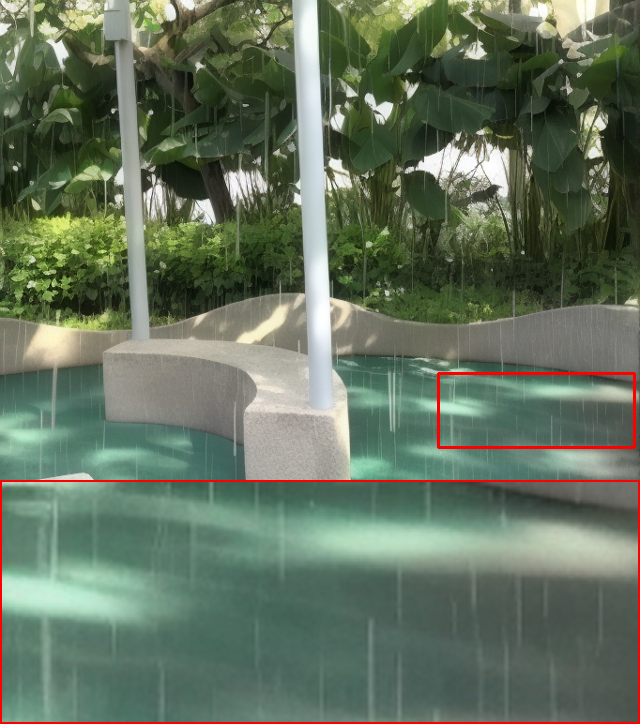}
&
\includegraphics[width=0.12\textwidth]{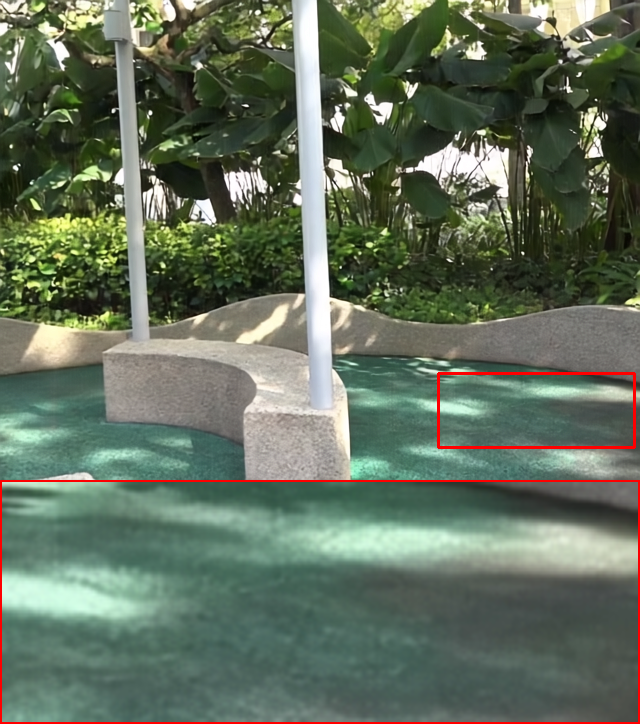}
&
\includegraphics[width=0.12\textwidth]{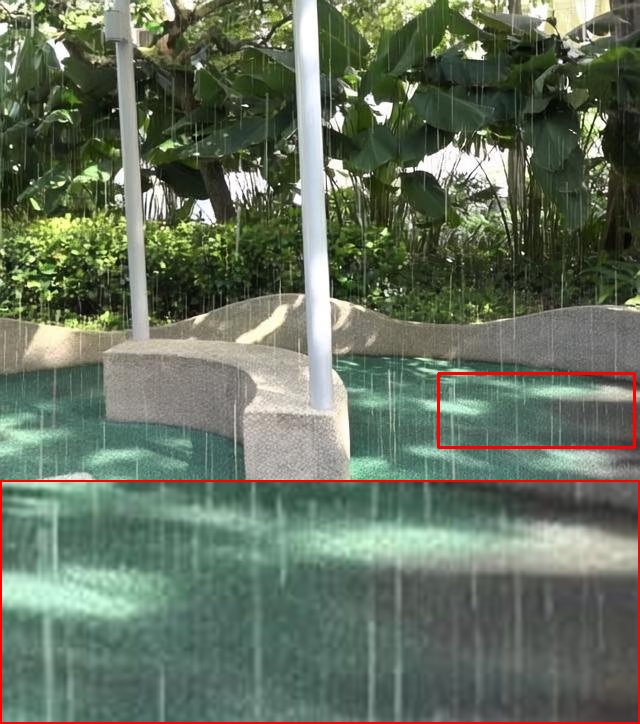}
&
\includegraphics[width=0.12\textwidth]{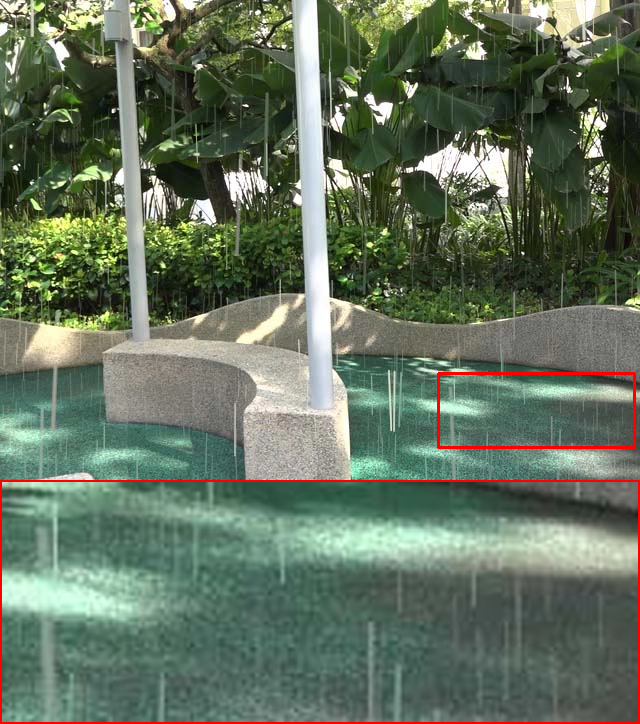}
&
\includegraphics[width=0.12\textwidth]{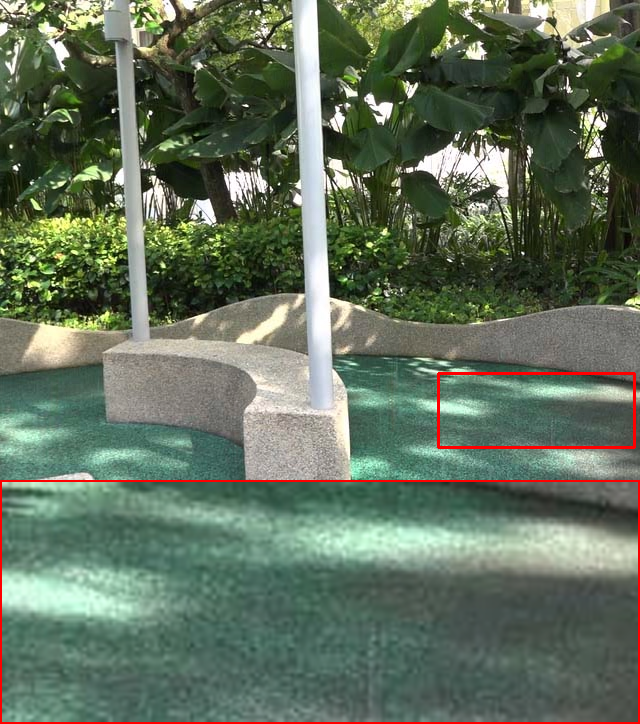}
&
\includegraphics[width=0.12\textwidth]{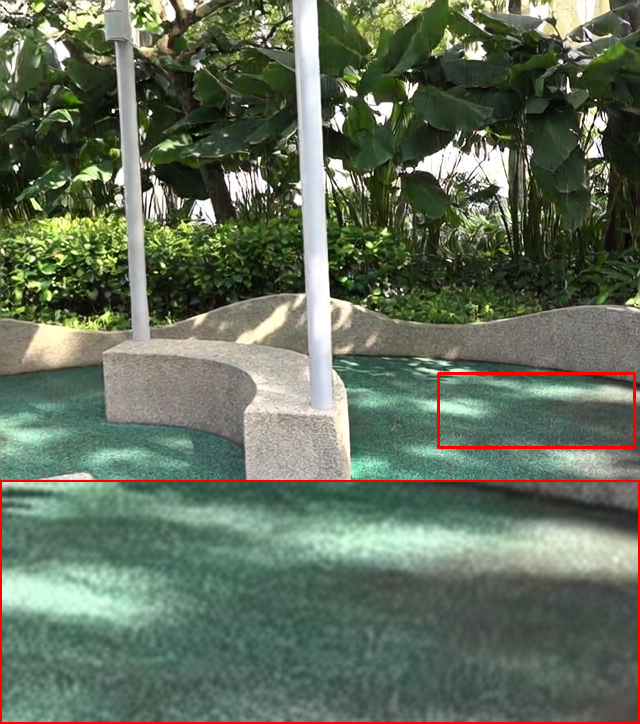}
&
\includegraphics[width=0.12\textwidth]{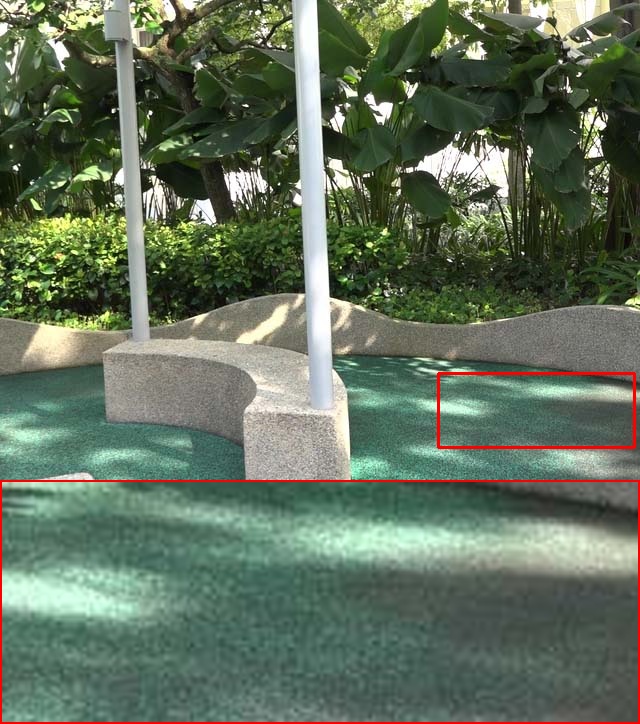}
\\

\includegraphics[width=0.12\textwidth]{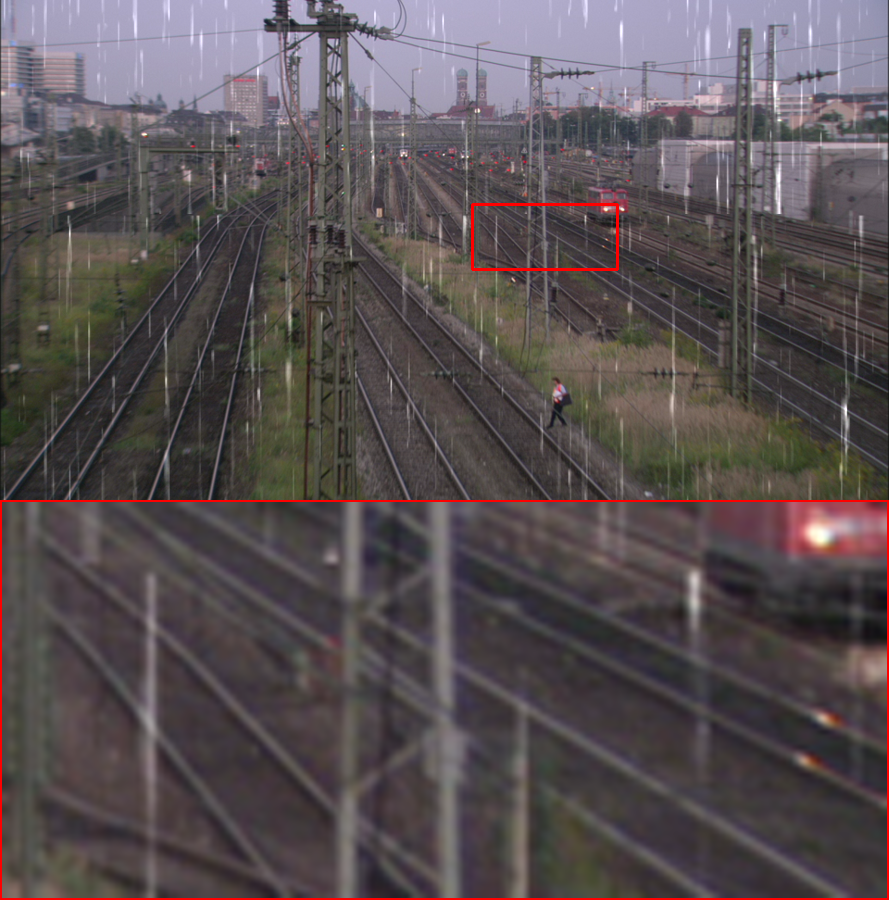}
&
\includegraphics[width=0.12\textwidth]{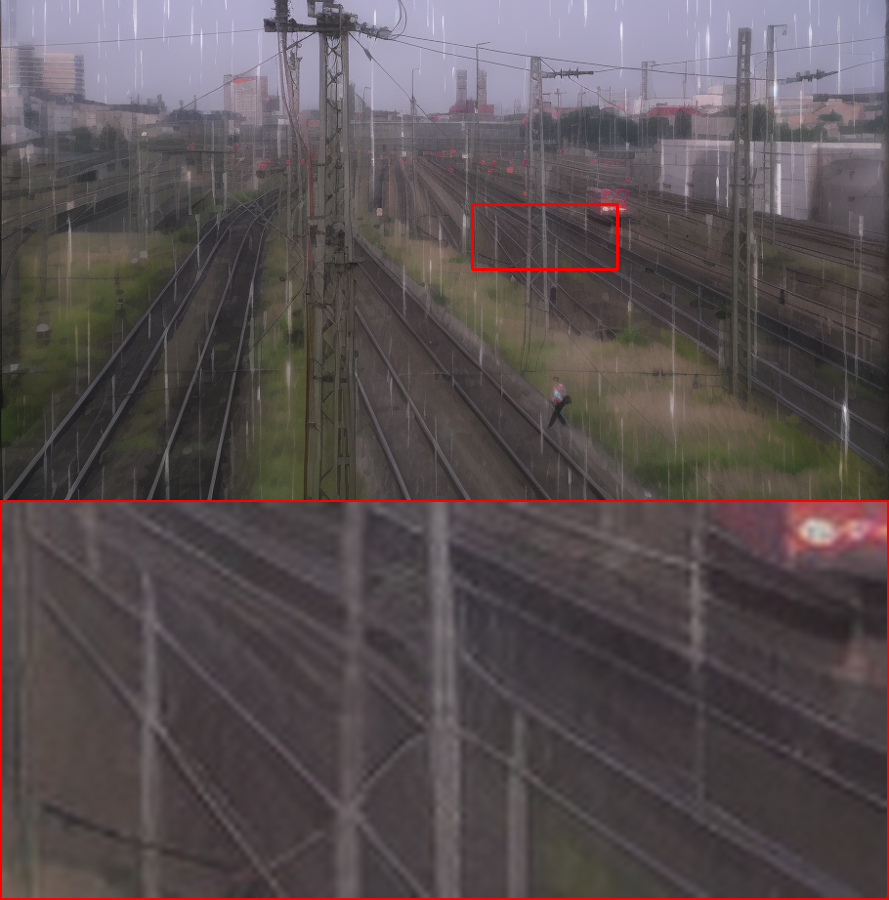}
&
\includegraphics[width=0.12\textwidth]{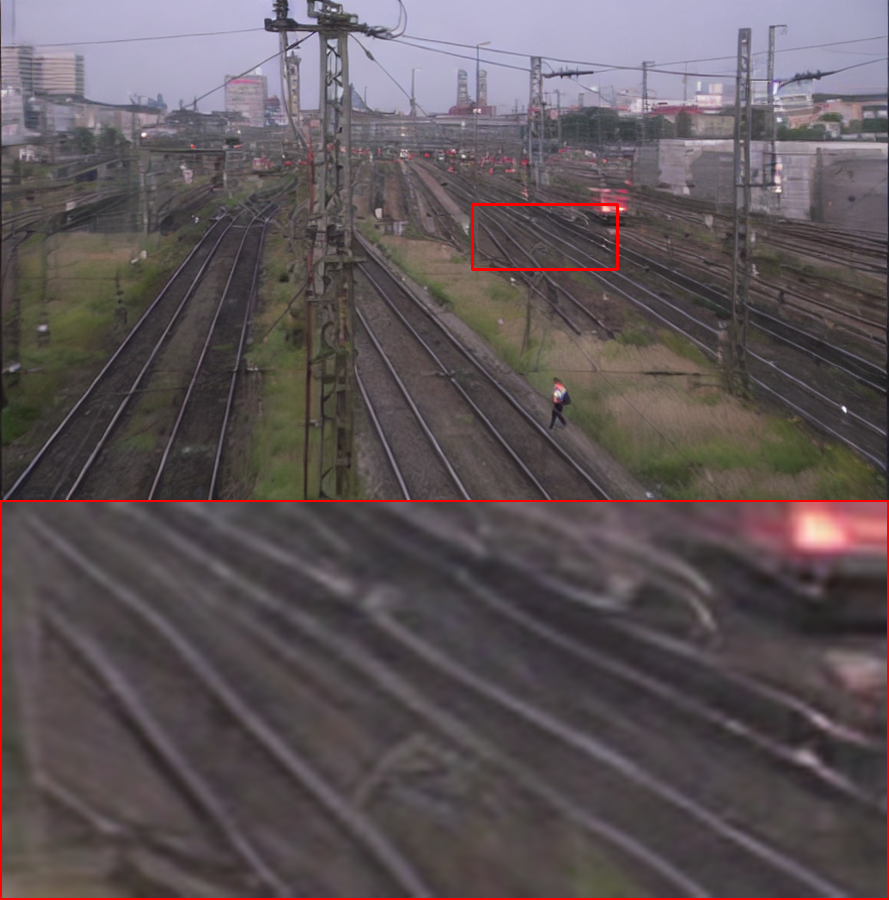}
&
\includegraphics[width=0.12\textwidth]{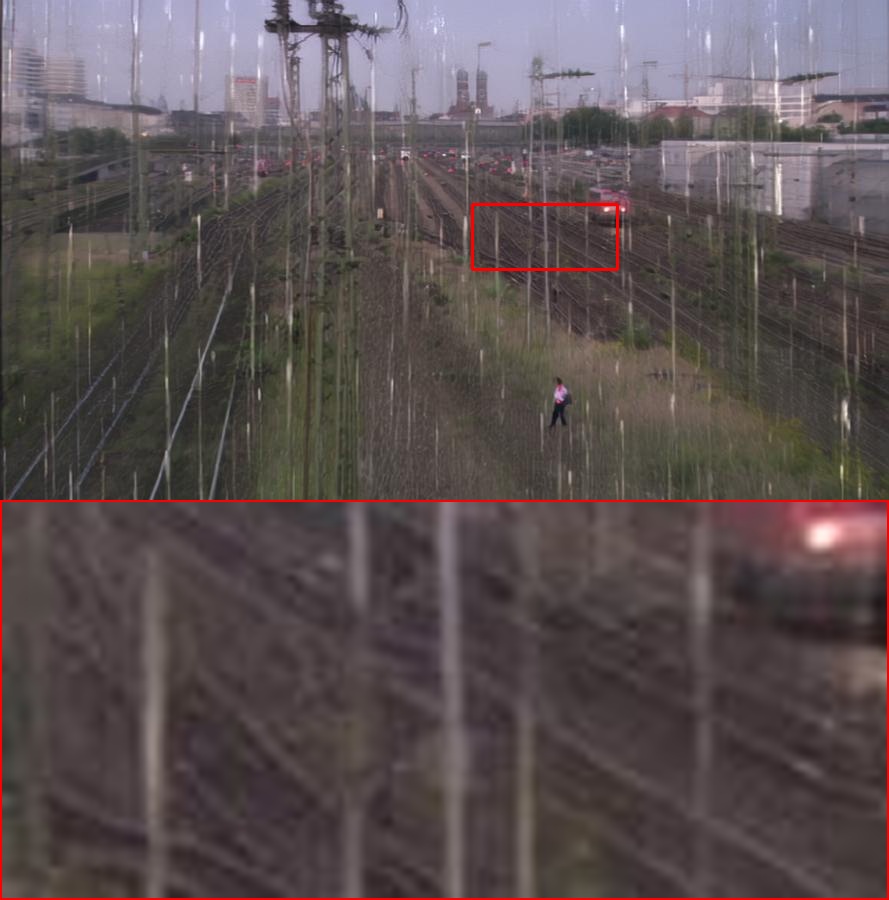}
&
\includegraphics[width=0.12\textwidth]{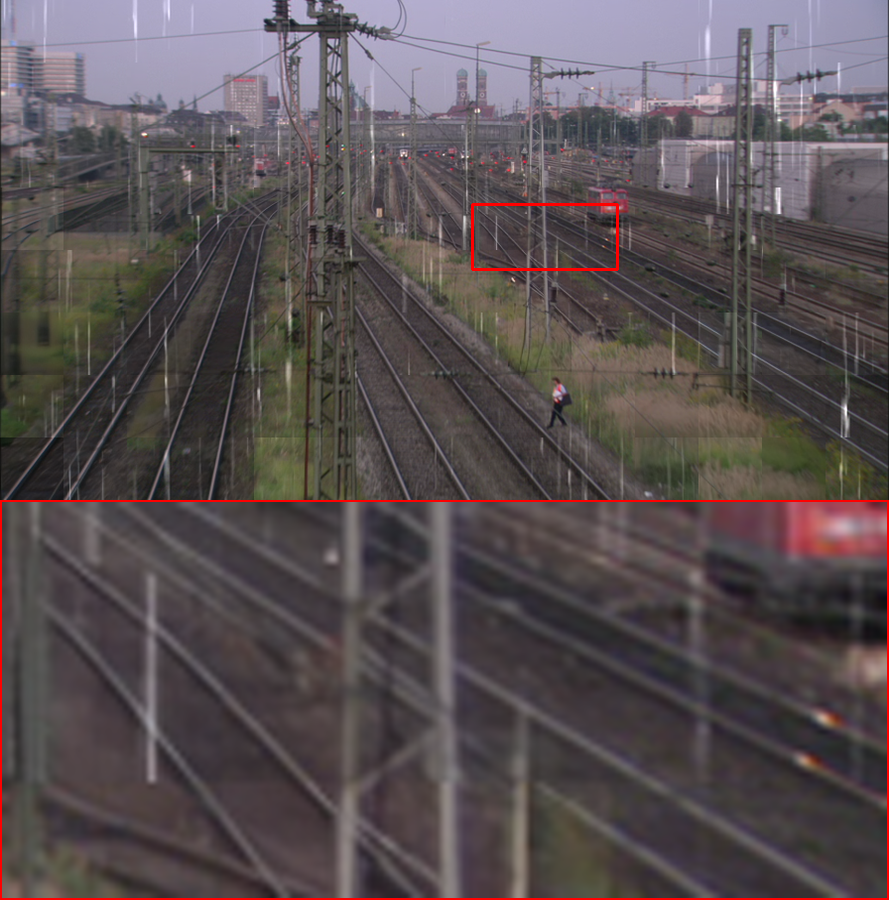}
&
\includegraphics[width=0.12\textwidth]{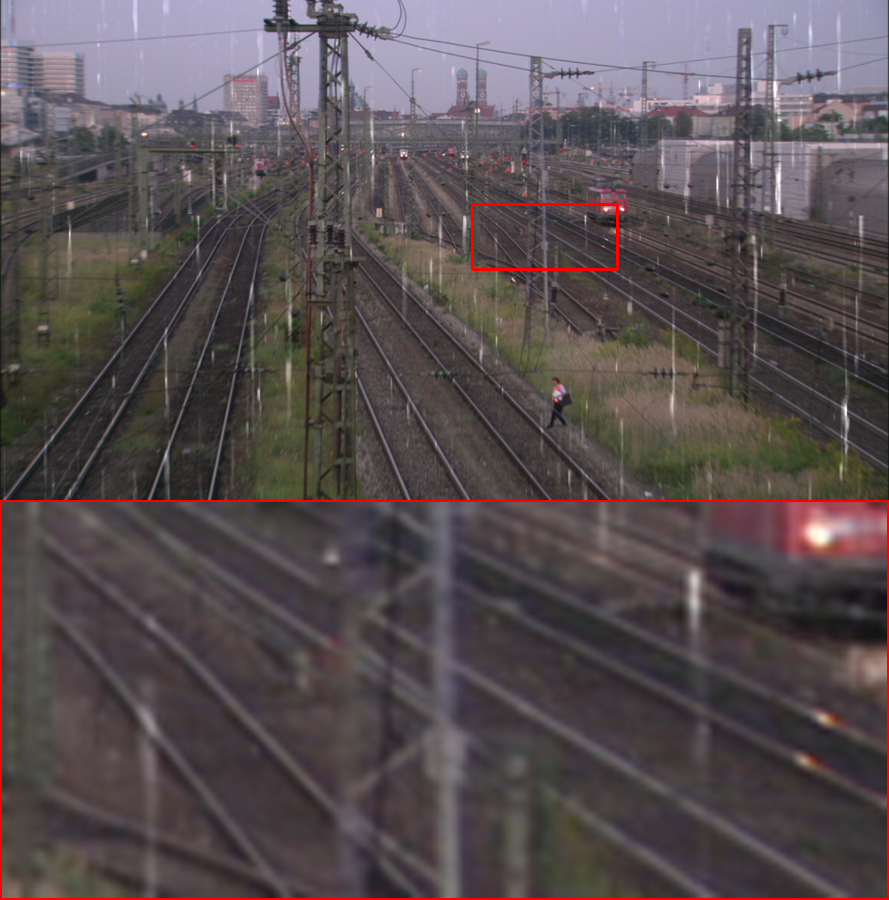}
&
\includegraphics[width=0.12\textwidth]{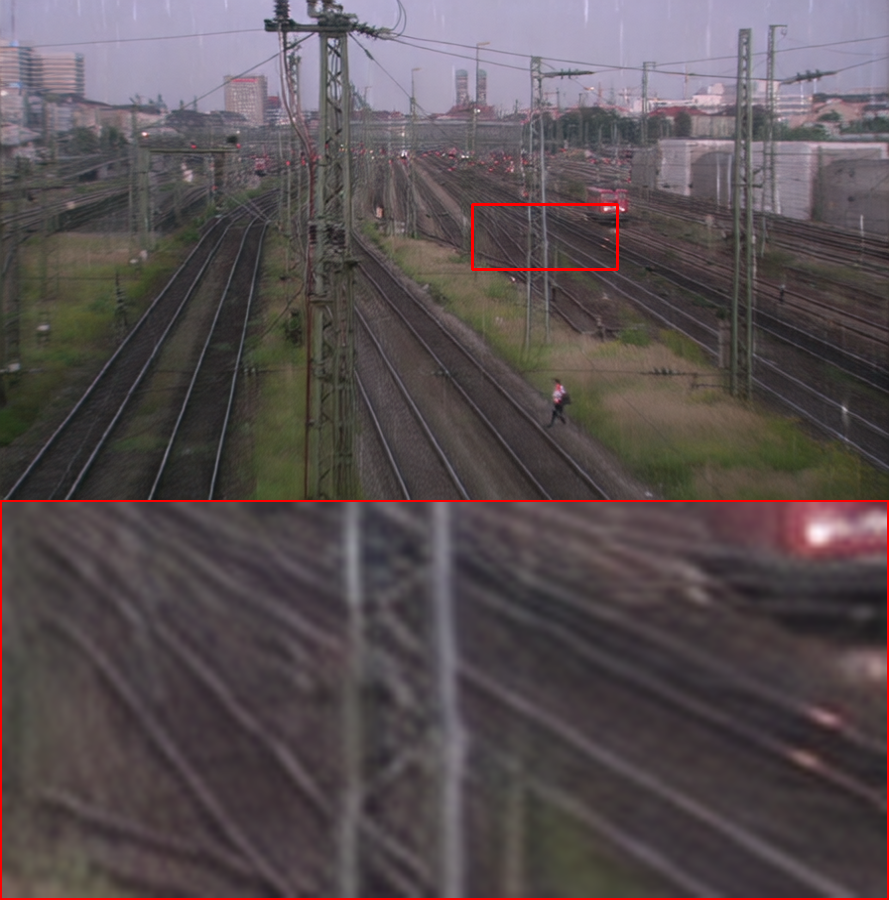}
&
\includegraphics[width=0.12\textwidth]{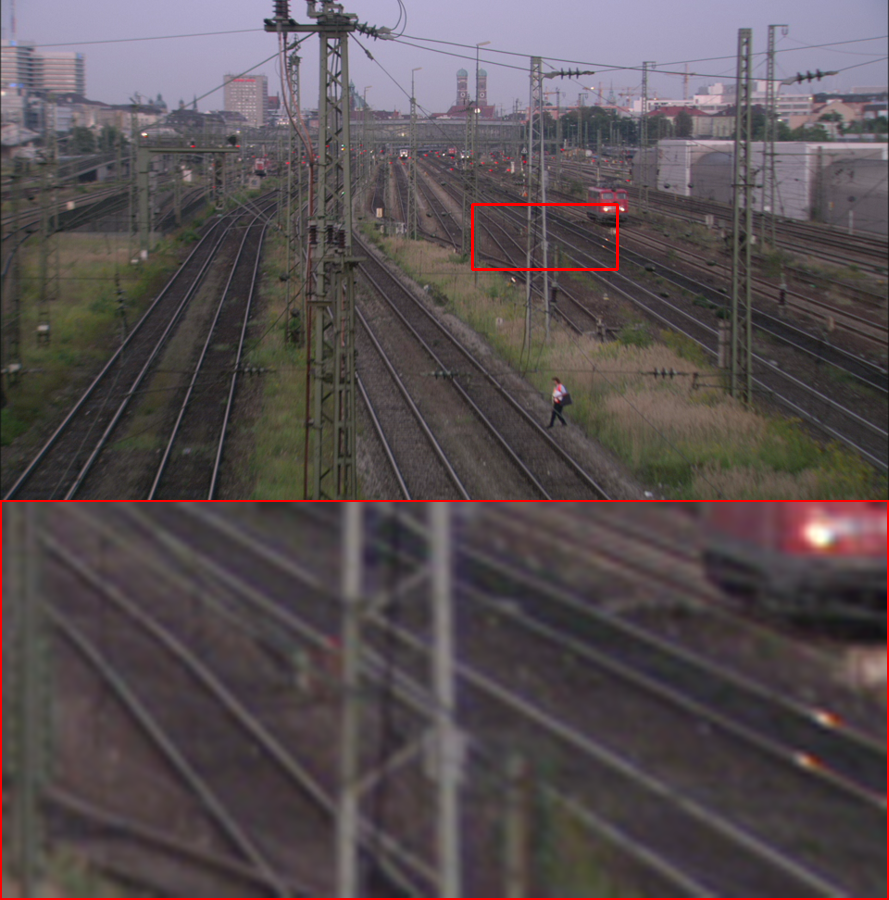}
\\

Rainy
&
DiffIR2VR
&
Diff-Plugin
&
ZSVD
&
WeatherDiff
&
T3-Diff
&
SGVD
&
GT
\\

\end{tabular}}

\caption{
Visual comparisons on synthetic video deraining datasets, including NTURain~\cite{chen2018robust} and RainSynLight25~\cite{liu2018d3r}.}\vspace{-4mm}
\label{fig:visual}
\end{figure*}

\begin{figure*}[!htb]
\centering
\captionsetup{font={scriptsize}}

\setlength{\tabcolsep}{1pt}

{\footnotesize
\begin{tabular}{cccccccc}

\includegraphics[width=0.12\textwidth]
{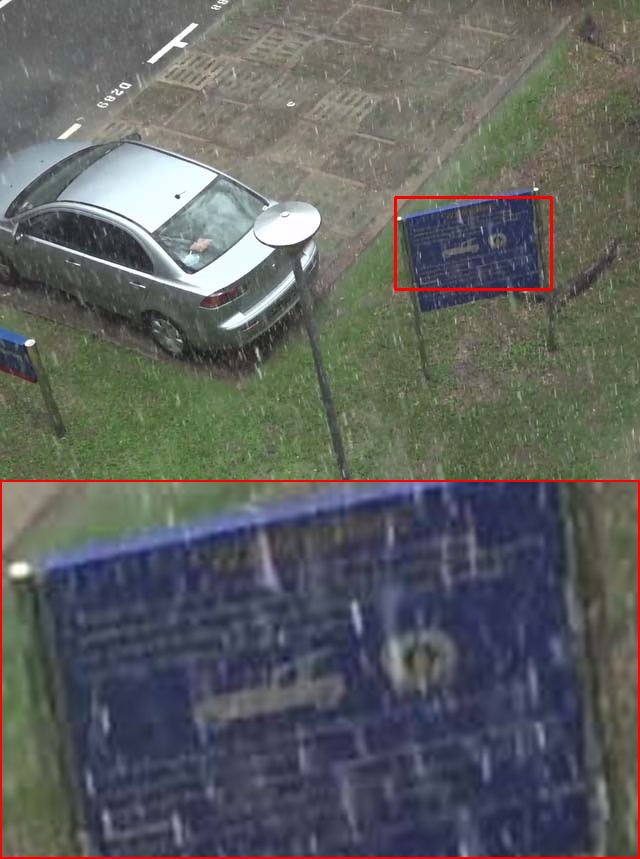}
&
\includegraphics[width=0.12\textwidth]
{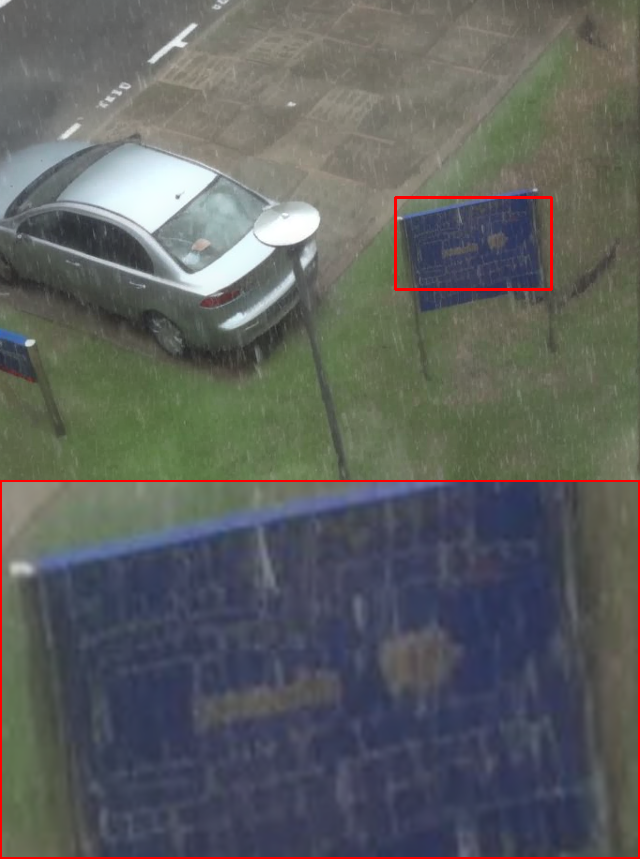}
&
\includegraphics[width=0.12\textwidth]
{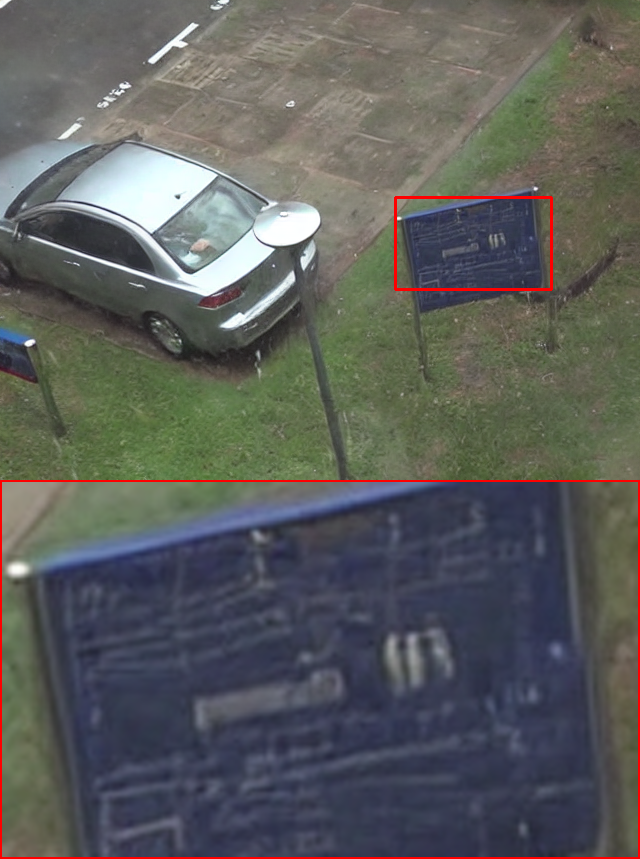}
&
\includegraphics[width=0.12\textwidth]
{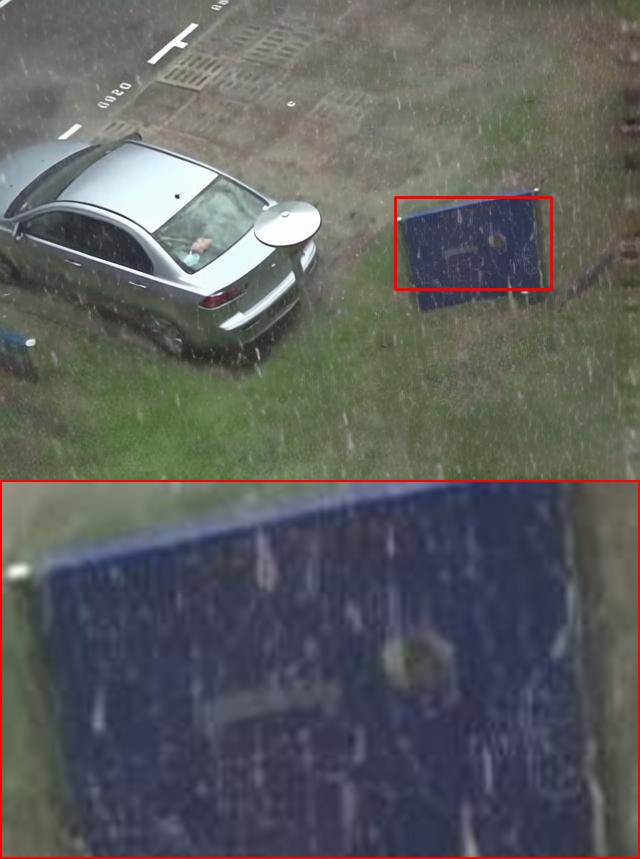}
&
\includegraphics[width=0.12\textwidth]
{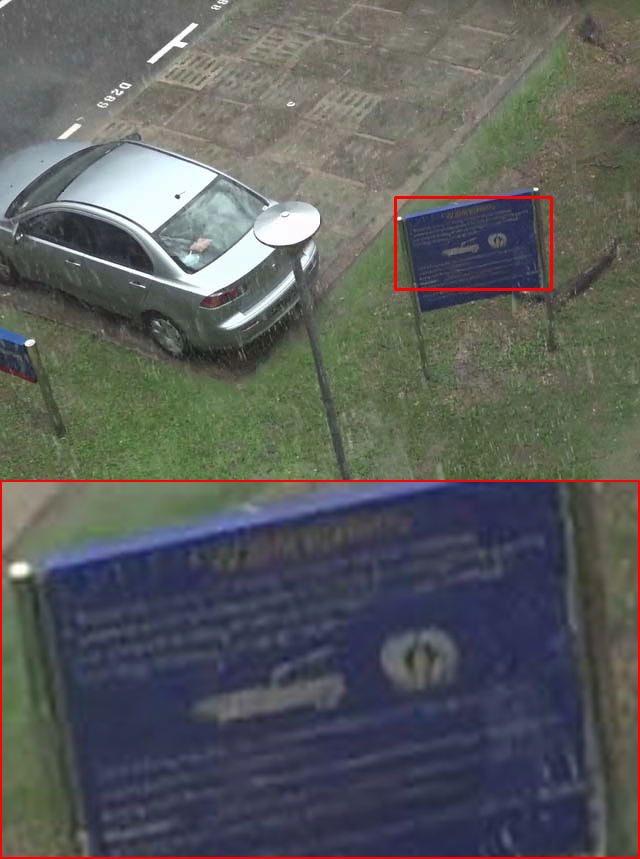}
&
\includegraphics[width=0.12\textwidth]
{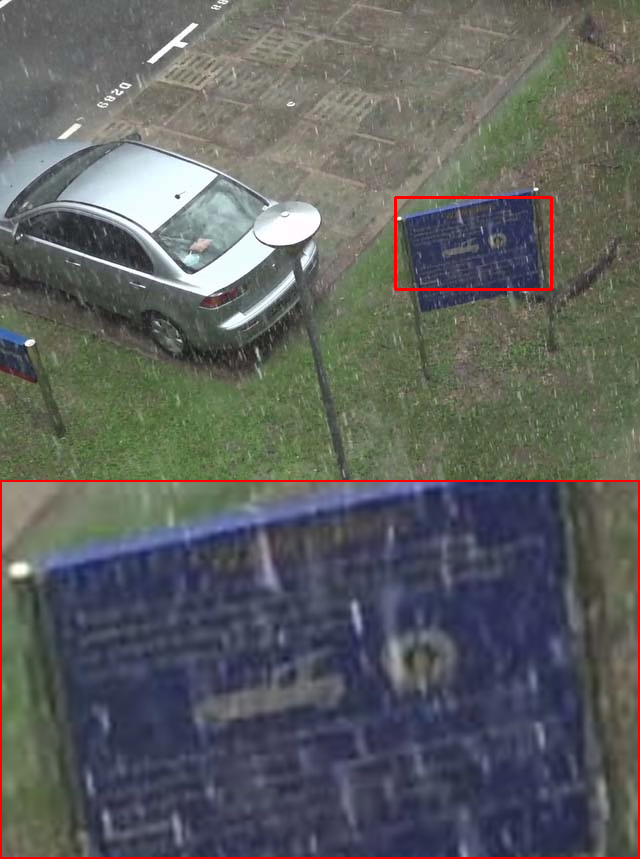}
&
\includegraphics[width=0.12\textwidth]
{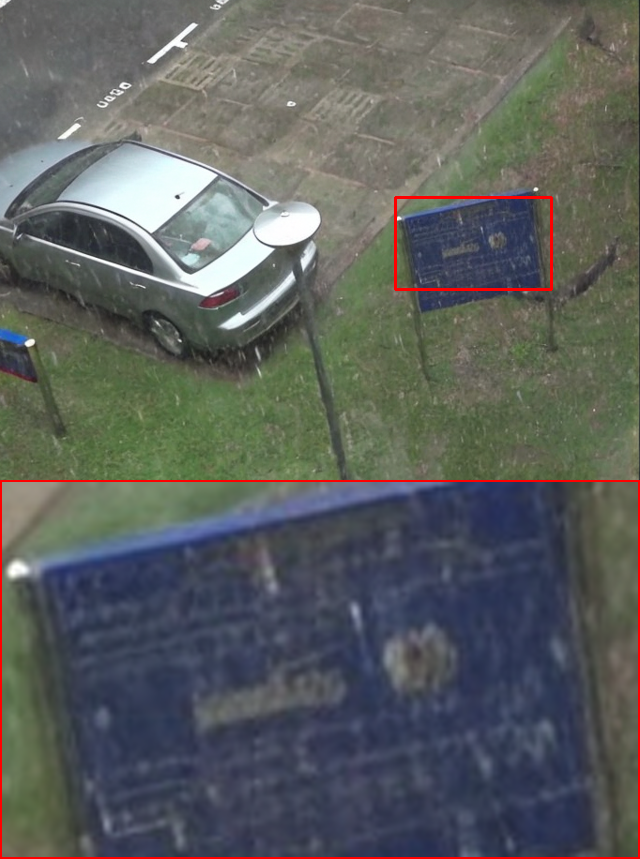}
&
\includegraphics[width=0.12\textwidth]
{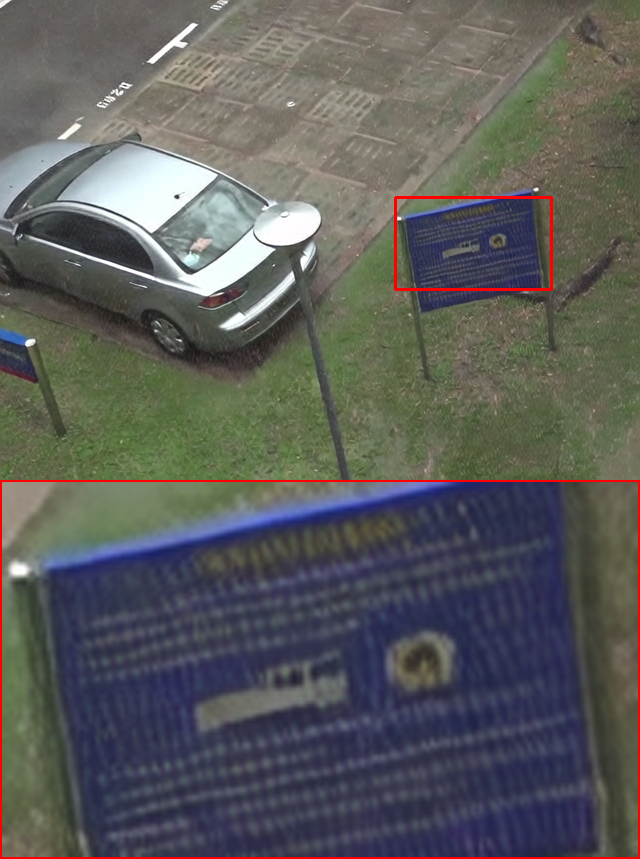}

\\

Rainy
&
DiffIR2VR
&
Diff-Plugin
&
ZSVD
&
WeatherDiff
&
T3-Diff
&
DiffBIR
&
SGVD

\end{tabular}}\vspace{-2mm}

\caption{
Visual comparisons on real-world rainy videos from the NTURain-Real dataset~\cite{chen2018robust}.
}\vspace{-4mm}
\label{fig:real_visual}

\end{figure*}

\noindent\textbf{Experimental Setup.}
SGVD is initialized from the official \href{https://huggingface.co/Wan-AI/Wan2.1-T2V-1.3B}{Wan-AI/Wan2.1-T2V-1.3B} release~\cite{wan2025wan}, with the pretrained DiT, VAE, and UMT5-XXL text encoder frozen. The complete model contains 7,298.47 M parameters, of which only the 71.67 M adapter parameters (0.98\%) are trainable. The shared training protocol uses 49-frame inputs, random $256\times256$ crops, AdamW with a learning rate of $10^{-3}$ and weight decay of $10^{-2}$, four gradient accumulation steps, and 100,000 iterations. ATC constructs each anchor by averaging predicted latent representations within a five-frame subclip. The text prompt \emph{Remove rain and restore clean background} is fixed throughout training and standard testing.\ifdefined\sgvdplatformnote\space\sgvdplatformnote\fi

\subsection{Comparison with State-of-the-Art Methods}

\noindent \textbf{Quantitative Comparisons.}
We compare SGVD with representative state-of-the-art restoration methods, including both diffusion-based and non-diffusion approaches. The diffusion-based baselines include DiffIR2VR~\cite{yeh2024diffir2vrzero}, Diff-Plugin~\cite{liu2024diffplugin}, DiffBIR~\cite{lin2023diffbir}, ZSVD~\cite{varanka2026zsvd}, DOVE~\cite{dove}, WeatherDiffusion (WeatherDiff)~\cite{weatherdiffusion}, and T3-Diff~\cite{chen2024t3diffweather}. The non-diffusion baselines include the Transformer-based DeLiVR~\cite{sun2026delivr} and the state-space-model-based RainMamba~\cite{wu2024rainmamba}.

As shown in Table~\ref{tab:sota}, SGVD achieves consistent improvements over the compared diffusion-based methods in reconstruction fidelity across different synthetic rain scenarios. In particular, SGVD obtains superior PSNR and SSIM performance with competitive NIQE scores on challenging rain conditions with severe occlusions, demonstrating its capability to recover missing contents while preserving reliable observations. SGVD also achieves better NIQE, CLIP-IQA, MUSIQ, and WE than DeLiVR and RainMamba on both NTURain-Syn and NTURain-Real. These advantages extend to cross-dataset evaluation, where SGVD leads the two non-diffusion baselines on all seven metrics on RainSynLight25 and five metrics on RainSynComplex25. Different from existing diffusion-based methods that apply generative priors globally, SGVD selectively activates generation according to restoration uncertainty.
This uncertainty-aware strategy balances fidelity preservation and perceptual recovery. Reliable regions maintain observation consistency, while ambiguous regions benefit from generative reconstruction. Retaining reliable observations can limit texture regeneration, reflecting the trade-off between pixel-level fidelity and perceptual similarity. SGVD balances these objectives to achieve strong reconstruction fidelity and competitive perceptual quality. Moreover, SGVD achieves highly competitive temporal consistency across the evaluated datasets and obtains the lowest warping error on NTURain-Syn and NTURain-Real. This improvement verifies that the proposed anchor-guided trajectory consistency mechanism effectively suppresses stochastic generation drift and maintains temporally coherent restoration.

\noindent\textbf{Qualitative Comparisons.}
Figure~\ref{fig:visual} presents visual comparisons under different rain degradation levels. Existing restoration methods generally remove rain artifacts but tend to generate over-smoothed results in severely corrupted regions due to insufficient prior information.
Meanwhile, global generative restoration methods can recover plausible structures but may introduce unrealistic details into reliable areas. In contrast, SGVD performs uncertainty-aware selective generation:
low-uncertainty regions preserve faithful observations, whereas high-uncertainty regions activate generative recovery.
As a result, SGVD produces cleaner backgrounds, sharper structures, and more realistic details under complex rain degradations.

\noindent \textbf{Real-World Evaluation.}
To further evaluate practical robustness, we conduct experiments on real-world rainy videos without clean references. As shown in Fig.~\ref{fig:real_visual}, SGVD generalizes well to diverse real-world rain conditions, including complex illumination changes and non-uniform rain patterns. Compared with existing methods, SGVD achieves better visual quality while avoiding excessive hallucination and preserving scene structures. These results demonstrate that uncertainty-aware selective generation provides a robust restoration strategy beyond synthetic benchmarks.

\subsection{Downstream Task Evaluation}

To further evaluate the practical utility of SGVD, we conduct downstream object detection and tracking experiments on the RVDT~\cite{sun2025semi} benchmark. The restored videos are directly fed into the downstream models without additional fine-tuning. For object detection, we employ YOLO-v3~\cite{redmon2018yolov3} and MEGA~\cite{chen2020mega}, while the tracking performance is evaluated using GTR~\cite{zhou2022gtr}. As shown in Table~\ref{tab:downstream}, SGVD achieves improved detection and tracking performance over rainy inputs. Specifically, SGVD obtains the highest mAP$_1$ of 42.02\%. Compared with the rainy input, SGVD improves mAP$_1$ and mAP$_2$ by 2.29 and 0.17 percentage points, respectively, demonstrating that the proposed selective generation strategy effectively restores discriminative object details. Furthermore, SGVD achieves the highest IDF1 of 66.4 and MOTA of 30.2, outperforming existing diffusion-based methods. This improvement verifies that the proposed ATC effectively reduces temporal generation drift and enhances cross-frame object association, leading to more reliable downstream video analysis.

\begin{table}[t]
\centering
\captionsetup{font={scriptsize}}
\caption{Object detection and tracking results on the RVDT benchmark. mAP$_1$ and mAP$_2$ denote the detection results obtained by YOLO-v3 and MEGA, respectively. The tracking results are obtained using GTR.}
\label{tab:downstream}
\setlength{\tabcolsep}{4pt}
\renewcommand{\arraystretch}{1.05}
\begin{tabular}{l|cc|ccc}
\toprule
\multirow{2}{*}{Method}
& \multicolumn{2}{c|}{Object Detection}
& \multicolumn{3}{c}{Object Tracking} \\
\cmidrule(lr){2-3}
\cmidrule(lr){4-6}
& mAP$_1\uparrow$
& mAP$_2\uparrow$
& IDF1$\uparrow$
& MOTA$\uparrow$
& MOTP$\uparrow$ \\
\midrule

Rainy Input
& 39.73
& 44.69
& 64.5
& 28.0
& 19.9 \\

ZSVD
& 38.86
& 43.32
& 64.8
& 26.4
& 20.3 \\

DOVE
& \underline{41.35}
& \textbf{45.87}
& 65.7
& \underline{30.1}
& 20.6 \\

WeatherDiff
& 37.82
& 42.12
& 63.5
& 29.5
& 18.8 \\

DiffBIR
& 38.06
& 44.56
& 65.3
& 27.9
& 20.4 \\

DiffIR2VR
& 41.08
& 43.35
& \underline{65.9}
& 28.1
& \textbf{21.0} \\

\textbf{SGVD}
& \textbf{42.02}
& \underline{44.86}
& \textbf{66.4}
& \textbf{30.2}
& \underline{20.8} \\

\bottomrule
\end{tabular}\vspace{-3mm}
\end{table}

\begin{figure}[t]
\centering
\includegraphics[width=\columnwidth]{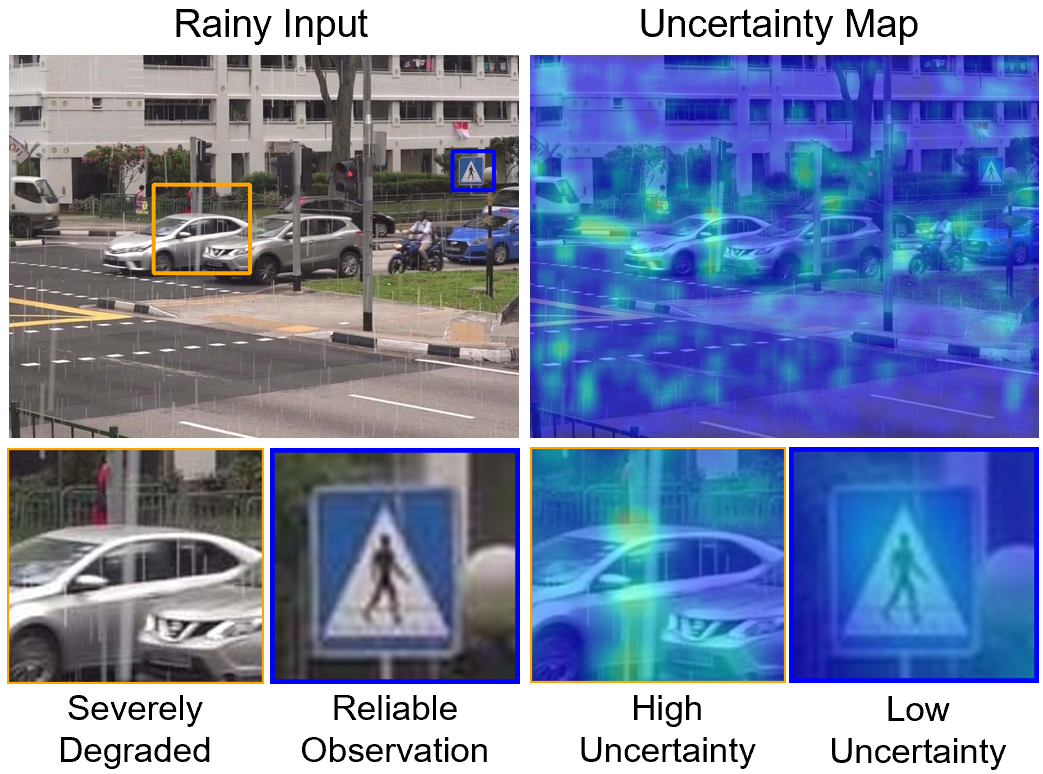}
\captionsetup{font={scriptsize}}
\caption{
Visualization of diffusion posterior uncertainty in SGVD. The uncertainty map is estimated from the discrepancy between two stochastic one-step diffusion predictions. Severely degraded regions exhibit stronger uncertainty responses, while reliable observations remain relatively low-uncertainty, indicating that the estimated uncertainty effectively reflects spatial variations in restoration ambiguity.
}\vspace{-2mm}
\label{fig:uncertainty}
\end{figure}

\begin{table}[htbp]
\centering
\captionsetup{font={scriptsize}}
\caption{Spearman correlation between the estimated uncertainty map and restoration error on synthetic video deraining datasets. Higher values indicate that regions with larger estimated uncertainty tend to exhibit larger restoration errors.}
\vspace{-2mm}
\label{tab:uncertainty_correlation}
\setlength{\tabcolsep}{4.5pt}
\renewcommand{\arraystretch}{1.05}
\resizebox{0.98\columnwidth}{!}{
\begin{tabular}{l|ccc}
\toprule
Metric
& NTURain-Syn
& RainSynLight25
& RainSynComplex25
\\
\midrule

Spearman Correlation ($\rho_s$)$\uparrow$
& 0.912
& 0.910
& 0.933
\\

\bottomrule
\end{tabular}}
\vspace{-3mm}
\end{table}

\subsection{Analysis of Diffusion Posterior Uncertainty}
\label{sec:uncertainty_analysis}

We further analyze whether the estimated uncertainty map accurately reflects restoration ambiguity. Figure~\ref{fig:uncertainty} presents uncertainty maps generated by the proposed diffusion posterior uncertainty estimation strategy. As observed, high uncertainty responses are mainly concentrated in regions with severe rain streaks, dense occlusions, and missing background structures. In contrast, reliable background regions exhibit relatively low uncertainty. This observation indicates that stochastic prediction diversity effectively captures spatial variations in restoration difficulty.

To quantitatively verify whether the estimated uncertainty corresponds to actual restoration ambiguity, we analyze the spatial correlation between uncertainty maps and reconstruction errors. Specifically, given the uncertainty map $M$ and the pixel-level restoration error
$E_i(p)=\|\hat{y}_i(p)-y_i(p)\|_1$, we employ Spearman's rank correlation coefficient $\rho_s$ to measure their monotonic spatial correspondence. Here $\hat{y}_i(p)$ and $y_i(p)$ denote the restored and ground-truth pixel values of frame $i$, respectively. A larger $\rho_s$ indicates that regions with higher estimated uncertainty tend to exhibit larger restoration errors. As shown in Table~\ref{tab:uncertainty_correlation}, consistently high correlations are observed across different synthetic datasets, showing that stochastic prediction disagreement tracks spatial variations in reconstruction error.

\begin{table}[htbp]
\centering
\captionsetup{font={scriptsize}}
\caption{Ablation study of the core components of SGVD on the NTURain-Syn test set. USG mainly improves fidelity and perceptual recovery, whereas ATC primarily improves temporal stability.}
\vspace{-2mm}
\label{tab:ablation_component}
\setlength{\tabcolsep}{2.2pt}
\renewcommand{\arraystretch}{1.05}
\resizebox{\columnwidth}{!}{
\begin{tabular}{c|cc|ccccccc}
\toprule
Model
& USG
& ATC
& PSNR$\uparrow$
& SSIM$\uparrow$
& LPIPS$\downarrow$
& NIQE$\downarrow$
& CLIP-IQA$\uparrow$
& MUSIQ$\uparrow$
& WE$\downarrow$
\\
\midrule

D1
&
&
& 23.0013
& 0.7819
& 0.1899
& 3.5131
& 0.3221
& 57.8211
& 0.0287
\\

D2
& \checkmark
&
& 26.2416
& 0.8732
& 0.1362
& 3.5114
& 0.3487
& 57.9213
& 0.0309
\\

D3
&
& \checkmark
& 25.9347
& 0.8638
& 0.1449
& 3.7052
& 0.3316
& 57.2845
& 0.0225
\\

\textbf{SGVD}
& \checkmark
& \checkmark
& \textbf{26.5174}
& \textbf{0.8810}
& \textbf{0.1310}
& \textbf{3.4224}
& \textbf{0.3565}
& \textbf{58.3749}
& \textbf{0.0207}
\\

\bottomrule
\end{tabular}}
\vspace{-2mm}
\end{table}

\subsection{Temporal Trajectory Consistency Analysis}
\label{sec:trajectory_analysis}

We evaluate whether the proposed Anchor-Guided Trajectory Consistency (ATC) effectively stabilizes the temporal evolution of one-step generative restoration.

\noindent\textbf{Quantitative Temporal Stability.} We first isolate the contribution of ATC using the component ablation results in Table~\ref{tab:ablation_component}.
The one-step restoration baseline without USG or ATC obtains a temporal warping error of $0.0287$.
Introducing USG alone substantially improves spatial restoration metrics, but only slightly changes the warping error from $0.0287$ to $0.0309$. This result indicates that uncertainty-guided spatial selection alone cannot fully resolve stochastic temporal drift.

\begin{figure}[!t]
\centering
\includegraphics[width=\columnwidth]{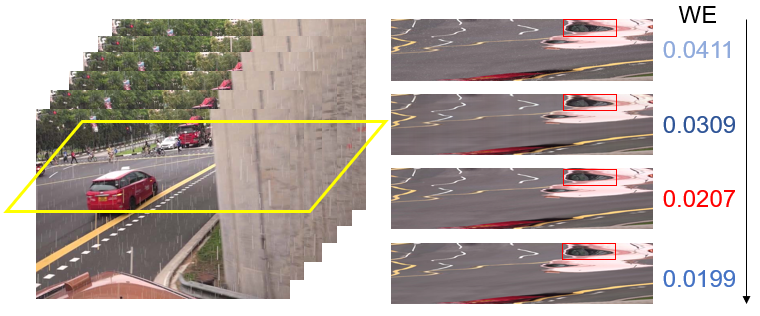}
\captionsetup{font={scriptsize}}
\caption{Temporal slice comparison of rainy input, SGVD without ATC, full SGVD, and ground truth (top to bottom). ATC improves structural consistency across frames and reduces temporal artifacts, yielding a lower \textbf{warping error (WE)}.
}\vspace{-3mm}
\label{fig:trajectory}
\end{figure}

In contrast, introducing ATC alone reduces the warping error from $0.0287$ to $0.0225$, while also improving PSNR and SSIM.
When USG and ATC are jointly incorporated, the complete SGVD further reduces the warping error to $0.0207$.
These results show that ATC improves temporal stability alongside reconstruction fidelity.
They also reveal the complementary relation: USG controls the spatial activation of generative priors, whereas ATC regularizes their temporal evolution.

\noindent\textbf{Temporal-Slice Visualization.} To qualitatively examine cross-frame stability, we construct temporal slices by extracting the same spatial line from consecutive frames and stacking them along the temporal dimension.
As shown in Fig.~\ref{fig:trajectory}, the result without ATC exhibits irregular fluctuations and discontinuous structural traces, particularly around object boundaries and texture-rich regions. These artifacts indicate that independently restored frames follow inconsistent stochastic trajectories.

With ATC, the temporal slices become substantially smoother and more continuous, while the underlying motion-induced variations remain visible. This observation confirms that ATC does not force neighboring frames to become identical.
Instead, it regularizes relative distances to the shared anchor within each five-frame subclip, allowing motion-dependent variations while discouraging incoherent temporal evolution.
Since ATC is used only during training, this temporal improvement introduces no additional inference overhead.

\noindent\textbf{Temporal Regularization Comparison.}
Table~\ref{tab:anchor_comparison} compares ATC with a neighboring-frame relative-trajectory loss while holding the backbone, USG, training budget, and loss-weight selection protocol fixed. The five-frame shared anchor reduces WE from 0.0323 to 0.0207 (35.9\%) and improves PSNR from 26.3309 to 26.5174 dB and SSIM from 0.8763 to 0.8810. NIQE and MUSIQ also improve to 3.4224 and 58.3749. The neighboring-frame objective obtains lower LPIPS and higher CLIP-IQA, while ATC offers stronger temporal consistency and fidelity with competitive perceptual quality. Relating each frame to the full five-frame subclip provides a common temporal reference beyond an immediate-neighbor pair.

\begin{table}[!t]
\centering
\captionsetup{font=scriptsize}
\caption{Temporal regularization comparison on NTURain-Syn. Best results are bold.}
\label{tab:anchor_comparison}
\footnotesize\setlength{\tabcolsep}{4pt}\renewcommand{\arraystretch}{1.10}
\begin{tabular*}{\columnwidth}{@{\extracolsep{\fill}}lrr@{}}\toprule
Metric & Neighboring-frame & Shared-anchor ATC\\\midrule
PSNR$\uparrow$ & 26.3309 & \textbf{26.5174}\\
SSIM$\uparrow$ & 0.8763 & \textbf{0.8810}\\
LPIPS$\downarrow$ & \textbf{0.1257} & 0.1310\\
NIQE$\downarrow$ & 3.5280 & \textbf{3.4224}\\
CLIP-IQA$\uparrow$ & \textbf{0.3671} & 0.3565\\
MUSIQ$\uparrow$ & 57.9113 & \textbf{58.3749}\\
WE$\downarrow$ & 0.0323 & \textbf{0.0207}\\
\bottomrule
\end{tabular*}
\end{table}

\subsection{Computational Efficiency}
\label{sec:efficiency}
We measure end-to-end inference on a single 48GB NVIDIA RTX 4090 GPU using 49-frame inputs at $640\times480$ resolution (width $\times$ height), batch size 1, and BF16 precision. After one warm-up run, latency is averaged over three timed runs of the same clip with CUDA synchronization. Model loading and disk I/O are excluded. Latency and FLOPs are normalized by the 49 output frames, and FPS is computed from the unrounded clip latency. FLOPs include matrix multiplications, convolutions, and attention, counting two operations per multiply-accumulate. The 30-step cost is calculated from the profiled fixed pipeline cost plus 30 repetitions of the per-step computation.

As shown in Table~\ref{tab:sampling_efficiency}, one-step SGVD requires 296 ms/frame and 21,548 G/frame, compared with 2,078 ms/frame and 226,611 G/frame for 30-step inference. This corresponds to approximately $7.0\times$ faster inference, an 86\% reduction in latency, and a 90.5\% reduction in FLOPs. By replacing iterative denoising with direct latent restoration, SGVD substantially reduces repeated backbone evaluations and improves video-processing throughput. The end-to-end speedup is smaller than the sampling-step ratio because operations such as VAE encoding and decoding incur costs independent of the number of restoration steps. The two uncertainty probes increase latency from 133 to 296 ms/frame and FLOPs from 11,372 to 21,548 G/frame, reducing throughput from approximately 7.52 to 3.37 FPS. These additional backbone evaluations provide spatial guidance for selective restoration, while reusing the frozen model without introducing trainable parameters. In addition, only 71.67 M of SGVD's 7,298.47 M parameters are trainable (0.98\%). Restricting optimization to the adapter reduces the number of parameters requiring stored gradients and optimizer states by approximately 99.02\%, enabling task-specific adaptation of the large pretrained model with a small parameter-update burden.

\begin{table}[!t]
\centering
\captionsetup{font=scriptsize}
\caption{End-to-end SGVD efficiency with and without uncertainty probing on 49-frame inputs at $640\times480$ resolution.}
\label{tab:sampling_efficiency}
\scriptsize\setlength{\tabcolsep}{2pt}\renewcommand{\arraystretch}{1.12}
\begin{tabular*}{\columnwidth}{@{\extracolsep{\fill}}lrrrrrr@{}}\toprule
Method & Steps & \makecell[r]{Time\\(ms/frame)} & FPS & \makecell[r]{FLOPs\\(G/frame)} & \makecell[r]{Total\\(M)} & \makecell[r]{Trainable\\(M)}\\\midrule
SGVD & 1 & 296 & 3.37 & 21548 & 7298.47 & 71.67\\
SGVD & 30 & 2078 & 0.48 & 226611 & 7298.47 & 71.67\\
SGVD w/o probes & 1 & 133 & 7.52 & 11372 & 7298.47 & 71.67\\
\bottomrule
\end{tabular*}
\end{table}

\subsection{Ablation Studies}
\label{sec:ablation}

After validating the Selective Generative Restoration paradigm and the estimated diffusion posterior uncertainty, we further conduct controlled experiments to isolate the contributions of individual components and analyze the sensitivity of the proposed framework.

\noindent\textbf{Effect of Core Components.}
Table~\ref{tab:ablation_component} evaluates the individual and joint effects of Uncertainty-Guided Selective Generation (USG) and Anchor-Guided Trajectory Consistency (ATC).
The baseline $\mathrm{D1}$ denotes the one-step generative restoration backbone without either component. Adding USG to the baseline improves PSNR from $23.0013$ to $26.2416$ and reduces LPIPS from $0.1899$ to $0.1362$.
It also improves the no-reference perceptual metrics, confirming that uncertainty-guided activation achieves a better balance between fidelity preservation and generative recovery.
However, its temporal warping error remains close to that of the baseline, indicating that spatial uncertainty modeling alone does not explicitly constrain stochastic temporal evolution. Adding ATC alone improves PSNR and SSIM and reduces the warping error from $0.0287$ to $0.0225$. These gains support the contribution of ATC to both spatial reconstruction and temporal stability. The complete SGVD achieves the best overall performance, reaching a PSNR of $26.5174$, an LPIPS of $0.1310$, and a warping error of $0.0207$. Compared with the baseline, these correspond to a PSNR improvement of $3.5161$ dB, an LPIPS reduction of $0.0589$, and a temporal warping error reduction of $27.9\%$. The consistent improvements across fidelity, perception, and temporal stability demonstrate that USG and ATC are complementary. USG determines where generation should be activated, while ATC determines how the activated generation should evolve consistently across frames.

\begin{table}[!htb]
\centering
\captionsetup{font={scriptsize}}
\caption{Effect of LQ latent utilization strategies on the NTURain~\cite{chen2018robust} synthetic dataset.}
\vspace{-2mm}
\label{tab:lq_utilization}
\setlength{\tabcolsep}{2.5pt}
\renewcommand{\arraystretch}{1.05}
\resizebox{\columnwidth}{!}{
\begin{tabular}{c|ccccccc}
\toprule
Variant
& PSNR$\uparrow$
& SSIM$\uparrow$
& LPIPS$\downarrow$
& NIQE$\downarrow$
& CLIP-IQA$\uparrow$
& MUSIQ$\uparrow$
& WE$\downarrow$
\\
\midrule

Start Only
& \underline{19.8903}
& 0.5467
& \underline{0.2796}
& \textbf{3.3830}
& \underline{0.3528}
& \underline{57.8211}
& 0.0256
\\

Condition Only
& 17.0639
& \underline{0.6489}
& 0.3435
& 3.7909
& 0.1627
& 42.4441
& \underline{0.0221}
\\

\textbf{Both (SGVD)}
& \textbf{26.5174}
& \textbf{0.8810}
& \textbf{0.1310}
& \underline{3.4224}
& \textbf{0.3565}
& \textbf{58.3749}
& \textbf{0.0207}
\\

\bottomrule
\end{tabular}}\vspace{-4mm}
\end{table}

\begin{table}[!htb]
\centering
\captionsetup{font={scriptsize}}
\caption{Effect of the number of stochastic posterior probes used for uncertainty estimation.
The two-probe configuration achieves the best overall accuracy--quality--efficiency trade-off.}
\vspace{-2mm}
\label{tab:ablation_uncertainty}
\setlength{\tabcolsep}{3.0pt}
\renewcommand{\arraystretch}{1.05}
\resizebox{0.94\columnwidth}{!}{
\begin{tabular}{c|ccccc}
\toprule
Number of Probes
& PSNR$\uparrow$
& LPIPS$\downarrow$
& NIQE$\downarrow$
& MUSIQ$\uparrow$
& WE$\downarrow$
\\
\midrule

w/o Uncertainty
& 25.6821
& 0.1478
& 3.8126
& 56.9821
& 0.0318
\\

Single Probe
& 26.0712
& 0.1404
& 3.6468
& 57.5316
& 0.0284
\\

Two Probes
& \textbf{26.5174}
& \textbf{0.1310}
& \textbf{3.4224}
& \textbf{58.3749}
& \underline{0.0207}
\\

Four Probes
& \underline{26.4938}
& \underline{0.1332}
& \underline{3.4517}
& \underline{58.1263}
& \textbf{0.0205}
\\

\bottomrule
\end{tabular}}
\vspace{-2mm}
\end{table}

\noindent\textbf{Effect of LQ Latent Utilization.}
We investigate the influence of different strategies for incorporating low-quality (LQ) latent information into the generative restoration process.
Specifically, we compare three variants: using the LQ latent only as the diffusion trajectory starting point, using it only as conditional guidance, and our full design that exploits the LQ latent for both trajectory initialization and conditioning.
As shown in Table~\ref{tab:lq_utilization}, using the LQ latent only as the starting point retains more information from the degraded observation, but provides insufficient guidance for recovering heavily corrupted contents, leading to limited fidelity and perceptual quality.
Using the LQ latent only as conditional guidance cannot adequately constrain the generative trajectory and results in substantial structural deviation from the target content.
By jointly using the LQ latent as both the trajectory starting state and the conditional signal, SGVD achieves substantially better reconstruction fidelity and perceptual quality, while also improving temporal consistency.
In particular, the full design achieves the best PSNR, SSIM, LPIPS, CLIP-IQA, MUSIQ, and WE, while maintaining competitive NIQE, demonstrating that the two roles of the LQ latent are complementary for balancing observation preservation and generative recovery.

\noindent\textbf{Number of Stochastic Posterior Probes.}
We next investigate the number of stochastic predictions used to estimate diffusion posterior uncertainty.
As shown in Table~\ref{tab:ablation_uncertainty}, removing uncertainty guidance degrades restoration performance, while a single prediction cannot explicitly characterize cross-sample posterior dispersion.
Using two independent predictions introduces an effective disagreement signal and achieves the best overall performance in PSNR, LPIPS, NIQE, and MUSIQ, indicating that pairwise stochastic diversity is sufficient for region-level uncertainty estimation.
Increasing the number of predictions from two to four only slightly improves WE from $0.0207$ to $0.0205$, while slightly degrading the other metrics and doubling the uncertainty-probing cost.
Therefore, we adopt two stochastic predictions as the default configuration for the best accuracy--quality--efficiency trade-off.

\begin{table}[!htbp]
\centering
\captionsetup{font={scriptsize}}
\caption{Sensitivity analysis of the weighting coefficients for flow matching, one-step consistency, and anchor-guided trajectory consistency.}
\vspace{-2mm}
\label{tab:ablation_loss}
\setlength{\tabcolsep}{3.0pt}
\renewcommand{\arraystretch}{1.05}
\resizebox{\columnwidth}{!}{
\begin{tabular}{c|ccc|ccccc}
\toprule
Variant
& $\lambda_{\mathrm{FM}}$
& $\lambda_{\mathrm{CM}}$
& $\lambda_{\mathrm{ATC}}$
& PSNR$\uparrow$
& SSIM$\uparrow$
& LPIPS$\downarrow$
& MUSIQ$\uparrow$
& WE$\downarrow$
\\
\midrule

Group1
& 1.0
& 1.0
& 0.05
& 26.2847
& 0.8746
& \underline{0.1328}
& 57.9482
& 0.0231
\\

Group2
& 1.0
& 1.0
& 0.10
& \textbf{26.5174}
& \textbf{0.8810}
& \textbf{0.1310}
& \textbf{58.3749}
& \textbf{0.0207}
\\

Group3
& 1.0
& 1.5
& 0.15
& \underline{26.4263}
& \underline{0.8789}
& 0.1358
& \underline{58.1637}
& 0.0220
\\

Group4
& 1.0
& 0.5
& 0.50
& 26.3518
& 0.8767
& 0.1341
& 58.0126
    & \underline{0.0214}
\\

\bottomrule
\end{tabular}}
\vspace{-2mm}
\end{table}

\noindent\textbf{Sensitivity to Loss Weights.}
We further analyze the sensitivity of SGVD to the weighting coefficients of the flow matching, one-step consistency, and anchor-guided trajectory consistency objectives. The corresponding results are presented in Table~\ref{tab:ablation_loss}. When the ATC loss weights are insufficient, the model cannot adequately balance generative trajectory learning, one-step endpoint consistency, and temporal trajectory regularization. The default configuration $\mathrm{Group2}$ achieves the best overall performance across PSNR, SSIM, LPIPS, MUSIQ, and temporal warping error, indicating a suitable balance among the three complementary objectives. Changing the weight combination in $\mathrm{Group3}$ and $\mathrm{Group4}$ does not yield further improvements, indicating sensitivity to the balance among the training objectives.%

\noindent\textbf{Text Conditioning.}
We compare the fixed training prompt \emph{Remove rain and restore clean background}, the shorter prompt \emph{Remove rain}, and an empty prompt at inference. The checkpoint, input clips, random seeds, sampling settings, guidance scale, and negative prompt are identical across the three settings. In Table~\ref{tab:prompt_comparison}, the original prompt leads on all seven metrics. Relative to the empty prompt, it improves PSNR by 0.1494 dB and SSIM by 0.0067, reduces LPIPS from 0.1368 to 0.1310, and lowers WE by approximately 1.9\%. The short and empty prompts give similar reconstruction quality, suggesting that visual conditioning supplies strong scene guidance, while the full task instruction provides a further improvement.

\begin{table}[!t]
\centering
\captionsetup{font=scriptsize}
\caption{Text-conditioning comparison on NTURain-Syn. Best results are bold.}
\label{tab:prompt_comparison}
\scriptsize\setlength{\tabcolsep}{1.5pt}\renewcommand{\arraystretch}{1.12}
\begin{tabular*}{\columnwidth}{@{\extracolsep{\fill}}lrrrrrrr@{}}\toprule
Prompt & PSNR$\uparrow$ & SSIM$\uparrow$ & LPIPS$\downarrow$ & NIQE$\downarrow$ & CLIP-IQA$\uparrow$ & MUSIQ$\uparrow$ & WE$\downarrow$\\\midrule
Original & \textbf{26.5174} & \textbf{0.8810} & \textbf{0.1310} & \textbf{3.4224} & \textbf{0.3565} & \textbf{58.3749} & \textbf{0.0207}\\
Short & 26.3683 & 0.8740 & 0.1359 & 3.4231 & 0.3433 & 57.9668 & 0.0211\\
Empty & 26.3680 & 0.8743 & 0.1368 & 3.4251 & 0.3467 & 58.2029 & 0.0211\\\bottomrule
\end{tabular*}
\end{table}

\noindent\textbf{Input-Length Generalization.}
We evaluate the same model trained on 49-frame inputs using 25-, 49-, and 81-frame inputs, without retraining or fine-tuning. Source sequences and scored frames are matched, and the resolution, prompt, and sampling settings are held fixed. In Table~\ref{tab:length_generalization}, the 81-frame configuration improves PSNR by 0.1365 dB and SSIM by 0.0087 over 49 frames and leads on five of the seven metrics. Its LPIPS of 0.1316 remains close to 0.1310 at 49 frames. The 25-frame setting retains comparable quality, with PSNR lower by 0.2217 dB. WE is 0.0214, 0.0207, and 0.0209 at 25, 49, and 81 frames, respectively, indicating stable temporal behavior on longer inputs. These results support length generalization over the evaluated range. The five-frame extent of the training regularizer does not fix the inference input length.

\begin{table}[!t]
\centering
\captionsetup{font=scriptsize}
\caption{Input-length evaluation using a model trained on 49-frame clips. Best results are bold.}
\label{tab:length_generalization}
\footnotesize\setlength{\tabcolsep}{4pt}\renewcommand{\arraystretch}{1.10}
\begin{tabular*}{\columnwidth}{@{\extracolsep{\fill}}lrrr@{}}\toprule
Metric & 25 frames & 49 frames & 81 frames\\\midrule
PSNR$\uparrow$ & 26.2957 & 26.5174 & \textbf{26.6539}\\
SSIM$\uparrow$ & 0.8784 & 0.8810 & \textbf{0.8897}\\
LPIPS$\downarrow$ & 0.1339 & \textbf{0.1310} & 0.1316\\
NIQE$\downarrow$ & 3.4480 & 3.4224 & \textbf{3.3489}\\
CLIP-IQA$\uparrow$ & 0.3497 & 0.3565 & \textbf{0.3609}\\
MUSIQ$\uparrow$ & 58.1333 & 58.3749 & \textbf{58.6926}\\
WE$\downarrow$ & 0.0214 & \textbf{0.0207} & 0.0209\\
\bottomrule
\end{tabular*}
\end{table}

\begin{figure}[!h]
\centering
\captionsetup{font=scriptsize}
{\scriptsize (a) Complex textures\par}\vspace{2pt}
\includegraphics[width=.323\linewidth]{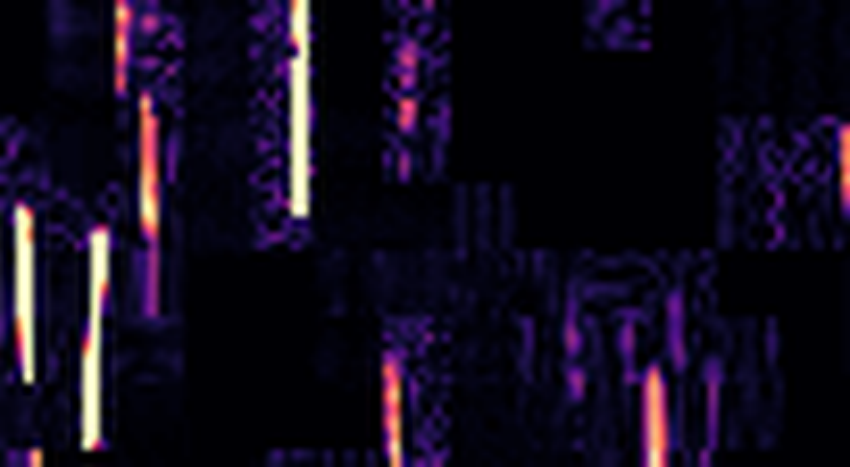}\hfill%
\includegraphics[width=.323\linewidth]{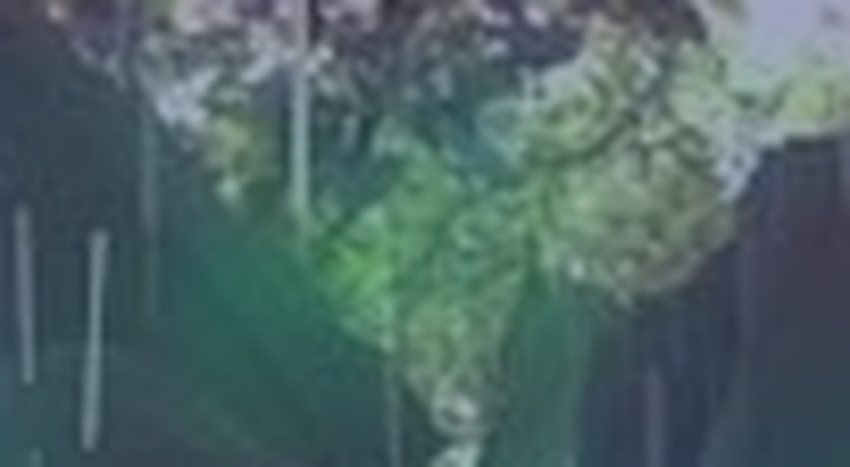}\hfill%
\includegraphics[width=.323\linewidth]{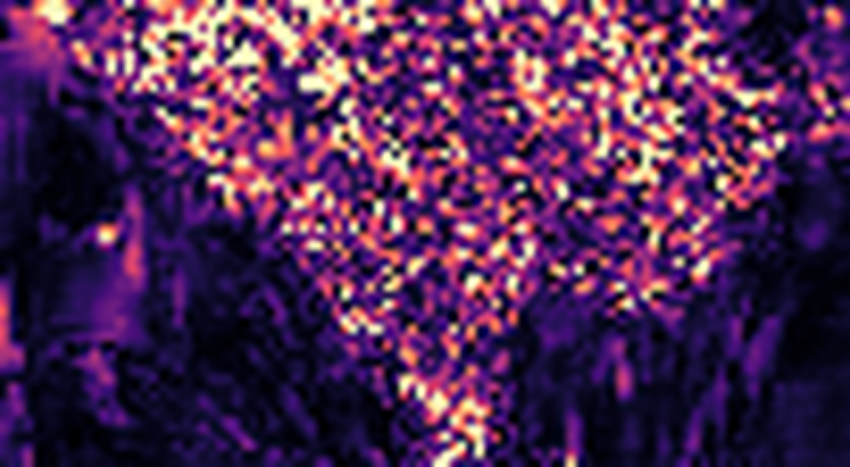}\par\vspace{5pt}
{\scriptsize (b) Motion boundaries\par}\vspace{2pt}
\includegraphics[width=.323\linewidth]{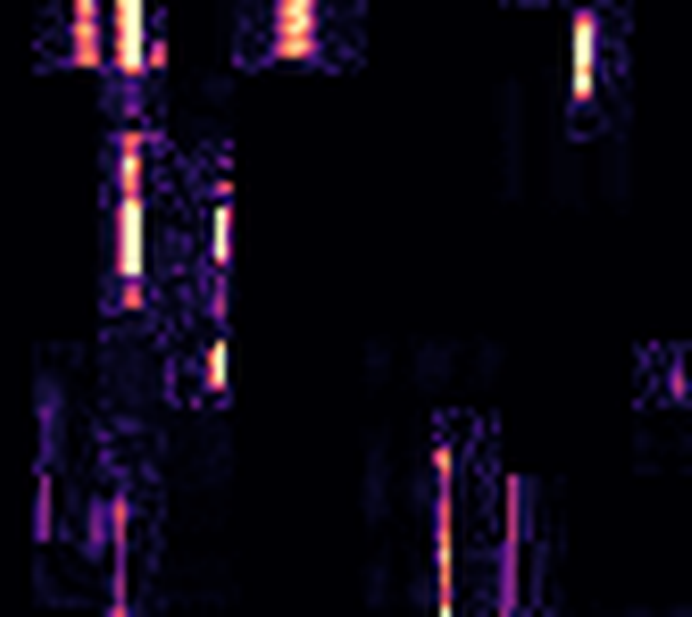}\hfill%
\includegraphics[width=.323\linewidth]{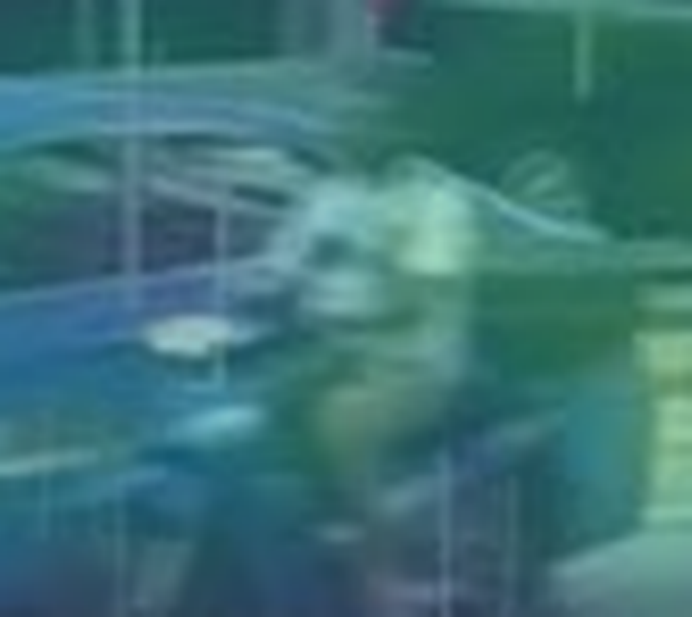}\hfill%
\includegraphics[width=.323\linewidth]{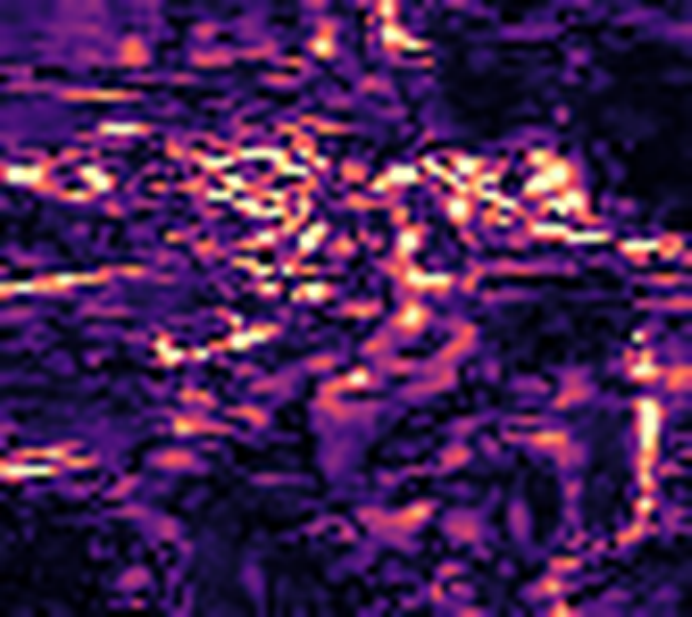}\par\vspace{5pt}
{\scriptsize (c) Illumination variations\par}\vspace{2pt}
\includegraphics[width=.323\linewidth]{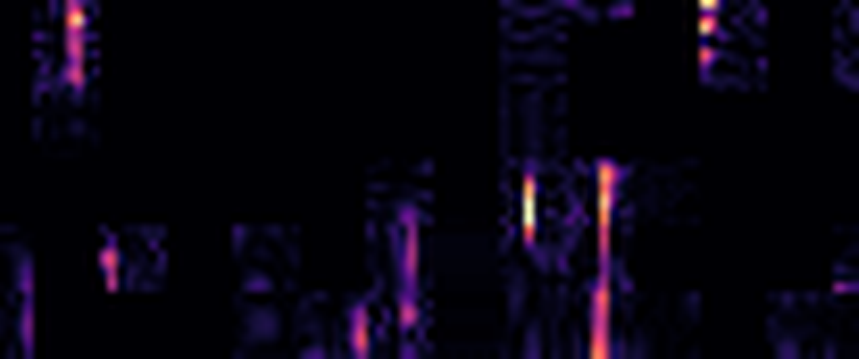}\hfill%
\includegraphics[width=.323\linewidth]{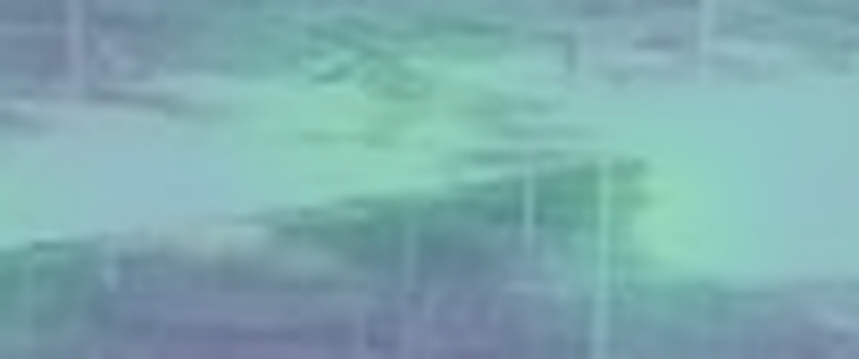}\hfill%
\includegraphics[width=.323\linewidth]{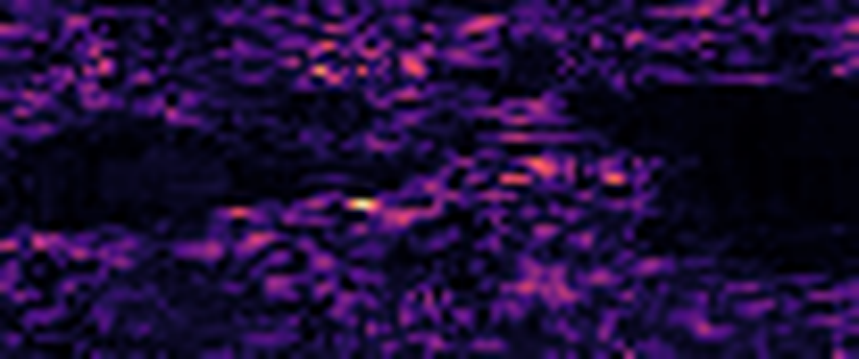}
\caption{Restoration examples in challenging scenes. From left to right: rain streak map, uncertainty overlay, and reconstruction error.}
\label{fig:uncertainty_challenging}
\end{figure}

\vspace{-2mm}
\subsection{Limitations}
\label{sec:limitations}
SGVD largely removes rain streaks in the challenging examples shown in Fig.~\ref{fig:uncertainty_challenging}, while remaining reconstruction errors are concentrated around complex textures, motion boundaries, and regions with varying illumination. Generative priors can recover plausible local details that differ from the clean reference. In addition, the two stochastic predictions used for uncertainty estimation add computational cost. Improving detail fidelity and reducing this computational cost are directions for future work.

\section{Conclusion}
\label{sec:conclusion}
In this paper, we revisit video deraining from the perspective of uncertainty-aware generative restoration and reveal that rain degradation is a spatially heterogeneous inverse problem.  Based on this insight, we propose \textbf{Selective Generative Video Deraining (SGVD)}, a new restoration framework that adaptively balances observation fidelity and generative inference.  SGVD exploits stochastic diversity of one-step generative predictions to estimate diffusion posterior uncertainty, enabling selective activation of generative priors in ambiguous regions while preserving reliable observations. Furthermore, an Anchor-Guided Trajectory Consistency mechanism is introduced to regularize stochastic restoration trajectories and improve temporal coherence without additional inference cost. Extensive experiments demonstrate that SGVD achieves strong reconstruction fidelity, competitive perceptual quality, and stable temporal behavior compared with existing video deraining methods.
This work provides a new perspective for diffusion-based restoration by shifting from global generation toward uncertainty-aware selective generation, which can be further extended to other restoration tasks with spatially varying degradation.

\bibliographystyle{IEEEtran}
\bibliography{final_manuscript/main}

\end{document}